\documentclass{article}

\usepackage[final]{neurips_2025}

\usepackage[utf8]{inputenc} 
\usepackage[T1]{fontenc}    
\usepackage{hyperref}       
\usepackage{url}            
\usepackage{booktabs}       
\usepackage{amsfonts}       
\usepackage{nicefrac}       
\usepackage{microtype}      
\usepackage{xcolor}         

\usepackage{enumitem}
\usepackage{amsmath}
\usepackage{amssymb}
\usepackage{amsthm}
\theoremstyle{definition}
\newtheorem{remark}{Remark}
\newtheorem{example}{Example}
\theoremstyle{plain}
\newtheorem*{theorem*}{Theorem}
\usepackage{dsfont}
\usepackage{algorithm}
\usepackage{algpseudocode}
\usepackage{paracol}
\usepackage{multicol}
\usepackage{hyperref}
\usepackage{cleveref}
\crefname{theorem}{Theorem}{Theorems}
\crefname{corollary}{Corollary}{Corollaries}
\crefname{lemma}{Lemma}{Lemmas}
\crefname{proposition}{Proposition}{Propositions}
\crefname{claim}{Claim}{Claims}
\crefname{remark}{Remark}{Remarks}
\crefname{section}{Section}{Sections}
\crefname{appendix}{Appendix}{Appendices}
\crefname{fact}{Fact}{Facts}
\crefname{algorithm}{Algorithm}{Algorithms}
\crefname{example}{Example}{Examples}
\crefname{table}{Table}{Tables}
\usepackage{abbreviations}
\usepackage{natbib}
\usepackage{graphicx}
\usepackage{subcaption}
\usepackage{multirow}
\usepackage{makecell}

\newcommand{\AlgName}{\texttt{INFEX}}
\newcommand{\Alg}{\mathsf{Alg}}
\newcommand{\Nopt}{N_{\text{opt}}}
\newcommand{\Nsub}{N_{\text{sub}}}
\newcommand{\Greed}[1]{\texttt{INFEX}(#1)}
\newcommand{\Tterm}{G(T)}
\newcommand{\constterm}[1]{G_{\text{const}}(#1)}
\newcommand{\consttermdefault}{\constterm{\tau_{\Alg}, f}}

\title{Infrequent Exploration in Linear Bandits}

\author{%
  Harin Lee \\
  University of Washington\\
  Seattle, WA, USA \\
  \texttt{leeharin@cs.washington.edu} \\
  \And
  Min-hwan Oh \\
  Seoul National University \\
  Seoul, South Korea \\
  \texttt{minoh@snu.ac.kr}
}

\begin{document}

\maketitle

\begin{abstract}    
We study the problem of infrequent exploration in linear bandits, addressing a significant yet overlooked gap between fully adaptive exploratory methods (e.g., UCB and Thompson Sampling), which explore potentially at every time step, and purely greedy approaches, which require stringent diversity assumptions to succeed. Continuous exploration can be impractical or unethical in safety-critical or costly domains, while purely greedy strategies typically fail without adequate contextual diversity. To bridge these extremes, we introduce a simple and practical framework, $\AlgName$, explicitly designed for infrequent exploration. $\AlgName$ executes a base exploratory policy according to a given schedule while predominantly choosing greedy actions in between. Despite its simplicity, our theoretical analysis demonstrates that $\AlgName$ achieves instance-dependent regret matching standard provably efficient algorithms, provided the exploration frequency exceeds a logarithmic threshold. Additionally, $\AlgName$ is a general, modular framework that allows seamless integration of any fully adaptive exploration method, enabling wide applicability and ease of adoption. By restricting intensive exploratory computations to infrequent intervals, our approach can also enhance computational efficiency. Empirical evaluations confirm our theoretical findings, showing state-of-the-art regret performance and runtime improvements over existing methods.
\end{abstract}

\section{Introduction}

\label{sec:introduction}

The multi-armed bandit (MAB) problem~\citep{lattimore2020bandit} captures a fundamental dilemma in sequential decision-making under uncertainty: at each time step, an agent must select an action (or arm) and receives feedback only from the chosen action, without observing the outcomes of alternative choices. Linear bandits generalize this problem by assuming that rewards follow a linear structure with respect to known arm features~\citep{abe1999associative, auer2002using, dani2008stochastic}, modeling diverse real-world scenarios such as clinical trials, recommendation systems, and adaptive pricing, where simultaneous learning and optimization are critical.

A central challenge in bandit settings is balancing \emph{exploration}—acquiring new information about uncertain arms—with \emph{exploitation}—leveraging existing knowledge to maximize immediate rewards. Classical algorithms, including Upper Confidence Bound (UCB)~\citep{auer2002finite,abbasi2011improved} and Thompson Sampling (TS)~\citep{thompson1933likelihood,agrawal2012analysis}, resolve this tension by exploring systematically at \emph{every time step}. These methods provide robust theoretical guarantees and strong empirical performance, forming the backbone of the MAB literature.

However, persistent exploration can be costly, risky, or ethically problematic in certain domains. For example, in healthcare or safety-critical settings, consistently experimenting with potentially suboptimal actions might lead to adverse or unacceptable outcomes. Consequently, it is often desirable to minimize exploration, performing it only when absolutely necessary. A straightforward alternative is the purely \emph{greedy policy}, which consistently selects the currently estimated optimal arm, offering simplicity and reduced risk by avoiding unnecessary experimentation.

Recent literature has studied conditions under which greedy algorithms achieve near-optimal performance in linear \emph{contextual} bandits~\citep{kannan2018smoothed,sivakumar2020structured, bastani2021mostly,raghavan2023greedy,kim2024local}. Crucially, these favorable theoretical guarantees of the greedy policy rely on strong distributional assumptions, such as sufficient \emph{diversity in observed contexts}, which naturally facilitates exploration. However, these guarantees fail to hold even in the standard linear bandit settings with fixed arm features, where the greedy approach typically incurs linear regret due to insufficient exploration and inadequate information acquisition (e.g., Example 1 in~\citet{jedor2021greedy}).

Thus, we are left with two extremes: at one extreme, greedy policies can succeed under strong diversity conditions (and may otherwise fail); at the other extreme, exploratory methods such as UCB or TS explicitly balance exploration and exploitation at every time step.
Surprisingly, there is a substantial gap between these extremes.
Specifically, the literature lacks rigorous studies on the impact of infrequent exploration on regret in linear bandit problems.
\footnote{While approaches such as the classic $\varepsilon$-greedy method---which introduces occasional stochastic exploration—and Explore-Then-Commit (ETC) algorithms---which perform an initial exploration phase followed by pure exploitation—are known to achieve suboptimal regret rates, in this work, we study whether infrequent exploration can be near-optimal.}

This raises fundamental open questions:
\begin{enumerate}
\item In order to achieve near-optimal performance (e.g., logarithmic regret), are we forced to explore at every time step, or can infrequent exploration suffice?
\item Can we devise an analytical framework to rigorously analyze methods with infrequent exploration, given that existing techniques may not directly apply?
\item How does the frequency of exploration affect regret performance?
\item Can infrequent exploration methods also demonstrate practical advantages beyond theoretical considerations?
\end{enumerate}

Answering these questions not only provides fundamental theoretical insights but also has significant practical implications, particularly in domains where frequent exploration carries substantial cost or risk. Moreover, even when exploration risks are low, thoroughly investigating these questions may still yield meaningful practical benefits, as exploration typically entails additional computational cost.

In this work, we rigorously address this critical gap by introducing a novel and practical framework, $\AlgName$ (\underline{Inf}requent \underline{Ex}ploration), designed explicitly for infrequent exploration in linear bandits. Given a base exploratory policy $\Alg$, our algorithm executes $\Alg$ according to a given schedule while predominantly making greedy action selections between these scheduled explorations. This hybrid approach naturally interpolates between fully exploratory and purely greedy strategies, offering fine-grained control over the exploration-exploitation trade-off. Notably, our approach is computationally efficient, which is particularly valuable in large-scale or real-time applications.

Our main contributions are summarized as follows:

\begin{itemize}

\item Our proposed framework $\AlgName$ is general and easily adoptable. It can seamlessly incorporate any (fully adaptive) linear bandit algorithm as the base policy, enabling broad applicability and straightforward integration into existing bandit implementations.

\item We analyze the regret of $\AlgName$ within the linear bandit framework. We show that despite interleaving greedy actions---which individually could incur linear regret in na\"{i}ve analysis---our algorithm achieves an instance-dependent regret matching that of $\texttt{LinUCB}$ (or $\texttt{OFUL}$)~\citep{abbasi2011improved}, provided the total number of exploratory time steps exceeds the order of $\log T$. This result demonstrates that the asymptotic regret behavior remains unaffected by the infrequency of exploration.
Furthermore, we complement the result by showing that the $\log T$ threshold is necessary (see \cref{thm:lower bound}).

\item We construct a new analytical framework for infrequent exploration that establishes regret bounds for $\AlgName$ with arbitrary exploration schedules.
Using this framework, we propose multiple exemplary exploration schedules and their resulting regret bounds.
The main distinction of our analysis comes from the observation that the estimation error of the optimal arm directly affects the regret, and we show that this error decreases as the number of optimal selections increases.

\item Furthermore, we derive a new instance-dependent regret bound for $\texttt{LinTS}$~\citep{agrawal2013thompson,abeille2017Linear}. This new theoretical insight may independently interest the broader bandit research community.

\item By limiting computationally intensive exploratory updates (e.g., posterior sampling or confidence set computations) to infrequent intervals, our algorithm significantly reduces runtime complexity compared to traditional approaches.

\item Empirical results, provided in \cref{sec:experiment}, substantiate our theoretical findings by demonstrating that, for suitable exploration schedules, $\AlgName$ outperforms both purely greedy and fully exploratory baselines in cumulative regret and computational efficiency.

\end{itemize}

\subsection{Related Work}

\paragraph{Full adaptive exploratory policies.} Classical bandit algorithms, such as Upper Confidence Bound (UCB)\citep{auer2002finite,abbasi2011improved} and Thompson Sampling (TS)\citep{thompson1933likelihood,agrawal2012analysis}, systematically balance exploration and exploitation at every time step. These approaches provide robust theoretical guarantees, including optimal logarithmic or sublinear regret bounds, and have been widely studied due to their effectiveness and simplicity. However, it remains an open question whether continuous exploration at every step is necessary or if infrequent exploration could suffice without compromising performance.

\paragraph{Greedy policies.} Recently, significant research has investigated conditions under which purely greedy algorithms achieve near-optimal performance, particularly within contextual bandit frameworks. Studies by~\citet{bastani2021mostly,kannan2018smoothed,sivakumar2020structured,oh2021sparsity,raghavan2023greedy,kim2024local} have shown that greedy policies can implicitly benefit from exploration when strong distributional assumptions, such as sufficient contextual diversity, are satisfied.
For instance, \citet{kannan2018smoothed,sivakumar2020structured,raghavan2023greedy} assume that the context vectors are perturbed by a multivariate Gaussian distribution at each time step, forcing the context distribution to be diverse.
\citet{kim2024local} study a more general class of distributions under which greedy policies achieve polylogarithmic regret.
While these findings identify specific scenarios favoring greedy methods, they leave unresolved how one should approach less ideal settings—such as linear bandit problems with fixed arm features lacking contextual diversity or stochastic variation, precisely the scenario addressed in our paper. In such standard linear bandit settings, purely greedy policies typically incur linear regret due to insufficient information gathering~\citep{jedor2021greedy}, highlighting the necessity of explicit exploration.

\paragraph{Randomized/scheduled forced exploration.} To incorporate explicit exploration in a simple manner, $\varepsilon$-greedy algorithms randomly explore arms with a small probability at each step~\citep{lattimore2020bandit,tirinzoni2022scalable}. While intuitive and computationally efficient, $\varepsilon$-greedy policies are theoretically known to incur suboptimal regret. Another approach, forced-sampling~\citep{goldenshluger2013linear, bastani2020online, lee2025lasso}, involves exploration at predetermined intervals. For instance,~\citet{goldenshluger2013linear} demonstrate that scheduled forced-sampling combined with greedy exploitation can achieve polylogarithmic regret under favorable context distributions.
Explore-Then-Commit (ETC) methods represent another scheduled exploration approach~\citep{langford2007epoch, abbasi2009forced, garivier2016explore,perchet2016batched,hao2020high}, separating exploration and exploitation into distinct phases. ETC algorithms initially perform extensive exploration to identify promising actions, after which they commit exclusively to exploiting the best-identified arm. Despite their simplicity and intuitive appeal, ETC methods typically result in suboptimal regret compared to fully adaptive exploration strategies such as UCB and TS.

\paragraph{Infrequent exploration.} To the best of our knowledge, approaches combining greedy exploitation with infrequent exploration have received limited attention, particularly in linear bandit contexts. 
One related work by~\citet{jin2023thompson} studies multi-armed bandits without features and proposes a hybrid method that randomly chooses between Thompson Sampling and greedy selections. Their results highlight the potential theoretical benefits of strategically interleaving exploration and exploitation. Nevertheless, extending this hybridization concept rigorously to linear bandits and establishing near-optimal regret guarantees remains an important open question.

Despite extensive research on adaptive exploration methods, greedy algorithms, and scheduled exploration, significant gaps remain in understanding how exploration frequency affects regret in linear bandits. Key questions include: Is continuous exploration necessary for near-optimal performance, and can infrequent exploration achieve similar guarantees? Current analytical frameworks primarily address frequent exploration, highlighting the need for rigorous approaches tailored specifically to infrequent exploration scenarios.

\section{Problem Setting}
\label{sec:preliminaries}

We consider the stochastic linear bandit problem.
The agent is presented with a finite arm set $\Xcal \subset \BB^d$ with $| \Xcal | = K$, where $\BB^d$ is the $d$-dimensional unit ball.
At each time step $t = 1, 2, \ldots$, the agent selects an arm $X_t \in \Xcal$ and receives a real-valued reward $Y_t = X_t^\top \theta^* + \eta_t$, where $\theta^* \in \RR^d$ is an unknown parameter vector and $\eta_t$ is zero-mean $\sigma$-subGaussian noise.\footnote{$\eta_t$ satisfies $\Expec [ \exp(s \eta_t) \mid X_1, Y_1, \ldots, X_t] \le \exp(s^2 \sigma^2 / 2)$ for all $s \in \RR$.}
We assume that $\| \theta^* \|_2 \le S$, and that this bound is known to the agent, where $\| \cdot \|_2$ denotes the $\ell_2$ norm.
The \textit{optimal arm} is the arm with the highest expected reward and is denoted by $x^* := \argmax_{x \in \Xcal} x^\top \theta^*$.
We assume that it is unique for simplicity.

A linear bandit algorithm $\Alg$ is one that (possibly randomly) selects $X_t$ based on the history $X_1, Y_1, \ldots, X_{t-1}, Y_{t-1}$.
The cumulative regret $\Rcal_{\Alg}(T)$ of an algorithm $\Alg$ over $T$ time steps is defined as follows:
\begin{equation*}
\Rcal_{\Alg}(T) := \sum_{t=1}^T \left( x^{*\top} \theta^* - X_t^\top \theta^* \right) .
\end{equation*}
The goal of the agent is to minimize the cumulative regret.
We primarily focus on instance-dependent regret, meaning that we study the growth of $\Rcal_{\Alg}(T)$ for a fixed problem instance $(\Xcal, \theta^*)$.

\section{Algorithmic Framework: \texorpdfstring{$\AlgName$}{INFEX}}
\label{sec:algorithm}

$\AlgName$ is a versatile and broadly applicable algorithmic framework designed for linear bandits and explicitly controls the frequency of exploration. The framework takes as input a base exploratory algorithm $\Alg$ and a predetermined exploration schedule $\Tcal_e$ (i.e., a set of time-step indices). At each time step in $\Tcal_e$, $\AlgName$ executes the exploratory algorithm $\Alg$, while at all other steps it acts greedily based on the ridge estimator. We denote the resulting hybrid algorithm as $\Greed{\Alg, \Tcal_e}$.

One notable advantage of $\AlgName$ is its generic design, enabling seamless integration of virtually any linear bandit algorithm as the exploratory component. This flexibility facilitates straightforward adaptation to various application domains and existing algorithmic frameworks. Furthermore, by clearly separating exploration and exploitation phases, $\AlgName$ achieves computational efficiency by limiting the frequency of computationally intensive exploratory procedures.

The pseudocode describing the procedure is provided in \cref{alg:MG}.

\begin{remark}[Substituting the ridge estimator.]
    The only properties of the ridge estimator used in our analysis are the boundedness of the online squared-loss regret,
    $\sum_{t=1}^T (X_t^{\top} \hat{\theta}_{t-1} - X_t^\top \theta^*)^2 = \Ocal(d^2 \log^2 T)$,
    and the fact that the estimation error
    $| x^\top \hat{\theta}_t - x^\top \theta^* |$
    decreases proportionally to $1 / \sqrt{n}$ when there are $n$ samples of $x$ in the data.
    Therefore, any estimator that satisfies similar properties may be used in place of the ridge estimator.
\end{remark}

\begin{algorithm}[t!] 
    \caption{$\Greed{\Alg, \Tcal_e}$: \underline{INF}requent \underline{EX}ploration} 
    \label{alg:MG}
    \begin{algorithmic}[1]
    \State Input : Base algorithm $\Alg$, exploration schedule $\Tcal_e \subset \NN$
    \State Initialize $V_0 = I_d$
        \For{$t=1, 2, ...,$}
            \If{$t \in \Tcal_e$}
                \State Choose $X_t$ according to $\Alg$ and observe $Y_t$
            \Else
                \State Compute ridge estimator $\hat{\theta}_{t-1} = V_{t-1}^{-1} \sum_{i=1}^{t-1} X_i Y_i$
                \State Choose $X_t = \argmax_{x \in \Xcal} x^\top \hat{\theta}_{t-1}$ and observe $Y_t$
            \EndIf
            \State Update $V_t = V_{t-1} + X_t X_t^\top$
        \EndFor
    \end{algorithmic}
\end{algorithm}

\section{Theoretical Analysis}

\subsection{Notations and Definitions}

Define $\reg_t := x^{*\top} \theta^* - X_t^\top \theta^*$ to be the instantaneous regret at time step $t$.
The main quantity that measures an instance’s difficulty is the \textit{minimum gap}, defined as
$\Delta := x^{*\top} \theta^* - \max_{x \in \Xcal \setminus \{ x^* \}} x^\top \theta^*$.
It represents the smallest possible non-zero instantaneous regret.

For two positive functions $f(x)$ and $g(x)$, we write $f(x) = \Ocal(g(x))$ if there exists a constant $C > 0$ such that $f(x) \le C g(x) + C$ for all $x$.
When $x$ is a positive real number and $ \lim_{x \rightarrow \infty} \frac{g(x)}{f(x)} = 0$, we write $f(x) = \omega(g(x))$.
In our analysis, we treat $d$, $T$, $K$, and $\Delta$ as variables, and regard all other quantities such as $\sigma$ and $S$ as constants.

We say an algorithm $\Alg$ attains (high-probability instance-dependent) \textit{polylogarithmic regret} if
$\Rcal_{\Alg}(T) = \Ocal \left( \frac{d^a}{\Delta^b} \log^c T \right)$ for some constants $a, b, c \ge 0$ with probability at least $1 - 1 / T$.
Note that our analysis holds for an arbitrary failure probability $\delta \in (0, 1]$.
For simplicity, we will mainly focus on the common choice $\delta \approx 1 / T$.
Such high-probability bounds that hold with probability at least $1 - 1/ T$
immediately imply comparable expected-regret bounds.

Let $f(t) := | \Tcal_e \cap \{1, 2, \ldots, t\} |$ be the number of time steps at which $\Alg$ is executed by $\Greed{\Alg, \Tcal_e}$ up to time step $t$.
Hence, $f(t)$ is the frequency of exploratory steps up to time $t$.
Let $f^{-1}(n) := \min\{ t \in \NN : f(t) \ge n \}$ be the time step at which $\Alg$ is executed for the $n$-th time.
One particular exploration schedule of interest is the periodic schedule that executes $\Alg$ at a fixed interval. 
For a positive integer $m$, let $m\mathbb{N} := \{m, 2m, 3m, \ldots\}$ denote the set of positive multiples of $m$. 
Then, the exploration schedule that executes $\Alg$ every $m$ time steps corresponds to $\Tcal_e = m\mathbb{N}$, and the resulting algorithm is denoted by $\Greed{\Alg, m\mathbb{N}}$.

Let $\Nopt(T) := \sum_{t=1}^T \ind\{ x^* = X_t \}$ denote the number of times the optimal arm is selected up to time step $T$.
We define
$\alpha_t := \log \frac{\det V_t}{\det V_0}$
and
$\beta_t(\delta) := \sigma \sqrt{\alpha_t + 2 \log (1 / \delta)} + S$,
which are key quantities in the analysis of many linear bandit algorithms~\citep{abbasi2011improved}.
For simplicity, we let $\beta_t := \beta_t( 1 / T)$ for all $t$.

\subsection{Main Results}

In this section, we analyze the regret bound of $\Greed{\Alg, \Tcal_e}$.

\begin{theorem}[Regret of $\AlgName$]
\label{thm:main theorem}
    Let $\Alg$ be a linear bandit algorithm that attains polylogarithmic regret, specifically $\Rcal_{\Alg}(T) = \Ocal \left( \frac{d^a}{\Delta^b} \log^c T \right)$ with probability at least $1 - 1 / T$ for some constants $a, b, c \ge 0$.
    Let $\Tcal_e \subset \NN$ be the set of exploratory time steps and $f(t) := | \Tcal_e \cap \{1, 2, \ldots, t\} |$ be the number of exploratory time steps up to time step $t$.
    Assume that $f(t) = \omega(\log t)$ as $t \rightarrow \infty$.
    Then, with probability at least $1 - 2 / T$, the regret of $\Greed{\Alg, \Tcal_e}$ is bounded as
    \begin{align*}
        \Rcal_{\Greed{\Alg, \Tcal_e}}(T) \le \Rcal_{\Alg}\left(f(T) \right) + \constterm{\tau_{\Alg}, f} + \Tterm \, ,
    \end{align*}
    where $\constterm{\tau_{\Alg}, f}$ is independent of $T$, $\tau_{\Alg} \in \NN$ is a constant determined by $\Alg$ satisfying $\tau_{\Alg} = \Ocal\left(\frac{d^a}{\Delta^{b+1}} \log^c \frac{d}{\Delta}\right)$, and
    \begin{align*}
        \Tterm = \Ocal\left( \frac{\left( \log T + d \log \log T + d \log \frac{d}{\Delta} \right)^2 }{\Delta} \right) \, .
    \end{align*}
    Bounds on $\constterm{\tau_{\Alg}, f}$ for some functions $f$ are provided in \cref{table:const term}.
\end{theorem}

\begin{table}[!ht]
\caption{Example bounds on $\constterm{\tau, f}$ for various functions $f$.
\textit{Epoch length} refers to the length between two consecutive executions of the base algorithm.}
\begin{center}
\begin{tabular}{l l l}
    \toprule
    \textbf{Example} of $f(t)$ & \textbf{Description} & $\constterm{\tau_{\Alg}, f}$\\
    \midrule
    $t / m$ & Epoch length is constant $m$ & $\Ocal \left( m \tau_{\Alg} + \frac{md}{\Delta} \log^2\frac{md}{\Delta} \right)$
    \\
    \midrule
    $t / (\log t)^r$ & Epoch length increases by $(\log t)^r$ & $\Ocal\left( \tau_{\Alg} \log ^r \tau_{\Alg} + \frac{d}{\Delta} \log^{2 + r} \frac{d}{\Delta} \right)$
    \\
    \midrule
    $t^r$ $(r \in (0, 1])$ & Epoch length increases by $t^{1 - r}$ & $\Ocal \left( \tau_{\Alg}^{1/r} + \frac{d^{1/r}}{\Delta^{2/r - 1}} \log^{2/r} \frac{d}{\Delta} \right)$
    \\
    \midrule
    $(\log t)^r$ $(r > 1)$ & Epoch length increases exponentially & $e^{\Ocal(\tau_{\Alg}^{1/r})} + \Delta e^{\Ocal( (d / \Delta^2)^{\frac{1}{r-1}})} $
    \\
    \bottomrule
\end{tabular}
\label{table:const term}
\end{center}
\end{table}

\paragraph{Discussion of Theorem~\ref{thm:main theorem}.}
In the regret bound of \cref{thm:main theorem}, only the terms $\Rcal_{\Alg}( f(T) )$ and $\Ocal\left( \frac{1}{\Delta} (\log T + d \log \log T)^2 \right)$ depend on $T$.
The first term corresponds to the regret of the base algorithm $\Alg$.
The second term bounds the additional regret incurred by the interleaved greedy selections, and it matches the instance-dependent bound of $\texttt{LinUCB}$~\citep{abbasi2011improved}.
We emphasize that these terms do not increase as the number of explorations decreases; in fact, the first term decreases.
Therefore, choosing a sparse exploration schedule does not worsen the asymptotic regret of $\Greed{\Alg, \Tcal_e}$, as long as it satisfies the condition $f(t) = \omega(\log t)$.
The trade-off from reduced exploration only appears in the constant term.
$\constterm{\tau_{\Alg}, f}$ is the cumulative regret incurred by the greedy selections for some initial time steps, where greedy selections do not have strong guarantees.
As shown in \cref{table:const term}, an excessively small number of explorations may result in exponential growth of the constant term with respect to $d / \Delta$, which may significantly degrade the algorithm's finite-time performance.
Meanwhile, exploration with constant periods or logarithmically growing epochs increases $\constterm{\tau_{\Alg}, f}$ only by a constant or a logarithmic factor.
For finite $T$, the least amount of exploration required to ensure that $\constterm{\tau_{\Alg}, f}$ does not exceed the order of $G(T)$ is determined by the relative magnitudes of $d$, $T$, and $\Delta$.
While it may be possible to allocate a minimal amount of exploration if all of these quantities are known, $\Delta$ is typically unknown to the agent, making it challenging to determine the optimal schedule.
In practice, we suggest that periodic or logarithmically growing epochs would be efficient.
However, it is very important to note that, even without knowing these quantities, $\AlgName$ achieves the same order of the regret compared to the vanilla fully adaptive exploration methods. 
In \cref{sec:experiment}, we demonstrate through numerical simulations that exploration with a fixed period of $5$ to $100$, so that $80\%$ to $99\%$ of the actions are greedy, yields favorable performance in terms of both regret and computational efficiency.

\paragraph{Obtaining minimax bound.}
We mainly focus on the instance-dependent bounds in this paper to show how the exploration schedule affects the regret for a fixed instance.
Meanwhile, providing the worst-case minimax regret bounds for infrequent exploration would also be an interesting problem.
While the asymptotic behavior of the instance-dependent bounds achieves the same order of polylogarithmic regret as long as the exploration number satisfies $\omega(\log t)$, we conjecture that this threshold would be too small to achieve the optimal $\Ocal(\sqrt{T})$ minimax guarantees.
Finding the optimal infrequent exploration strategy and trade-offs for the minimax regret bound would be an interesting open problem.

As an instantiation of $\AlgName$, we can choose $\Alg = \texttt{LinUCB}$~\citep{abbasi2011improved} or $\Alg = \texttt{LinTS}$~\citep{abeille2017Linear}, which are representative linear bandit algorithms.
To show that \cref{thm:main theorem} applies to both algorithms, we present their instance-dependent polylogarithmic regret bounds.
To the best of our knowledge, the instance-dependent bound for $\texttt{LinTS}$ is explicitly shown for the first time.
The proof of \cref{thm:lints} is deferred to \cref{appx:proof lints}.

\begin{theorem}
\label{thm:lints}
    $\texttt{LinTS}$~\citep{abeille2017Linear} achieves the following instance-dependent bound with probability at least $1 - \delta$:
    \begin{align*}
        \Rcal_{\texttt{LinTS}}(T) = \Ocal \left( \frac{ \min\{d\log \frac{dT}{\delta}, \log \frac{KT}{\delta} \} \left(\alpha_T + \log \frac{1}{\delta} \right)^2}{\Delta} \right)
        \, ,
    \end{align*}
    where $\displaystyle \alpha_T = \Ocal \left( \min \left\{ d \log T,  \log T + d \log \log T + d \log \frac{d}{\Delta} \right\} \right)$.
\end{theorem}

Furthermore, Theorem 5 in~\citet{abbasi2011improved} states that the regret of $\texttt{LinUCB}$ is $\Rcal_{\texttt{LinUCB}}(T) = \Ocal( \alpha_T^2 / \Delta)$ with the same bound on $\alpha_T$ as in \cref{thm:lints}.
Then, combined with the result of \cref{thm:main theorem}, we obtain the regret bounds for specific base algorithms.
We show some example regret bounds for $\AlgName$ when $\Alg$ is $\texttt{LinUCB}$ or $\texttt{LinTS}$ with varying exploration schedule in \cref{table:total regret}.
It demonstrates that the regret of $\AlgName$, instantiated with $\texttt{LinUCB}$ or $\texttt{LinTS}$, matches the regret bounds of the corresponding algorithms without infrequent exploration, up to factors independent of $T$.

\begin{table}[!ht]
\caption{Example regret bounds of $\AlgName(\Alg, \Tcal_e)$ with specific instantiations of $\Alg$ and $f(t)$. Each column shows the regret corresponding to each base algorithm. The final regret bound is the sum of the regret shown in the base regret row and the constant regret shown in the row with the corresponding exploration schedule.}
\begin{center}
\begin{tabular}{l  l l}
    \toprule
\multirow[b]{2}{*}{\makecell[l]{\textbf{Frequency of}\\\textbf{exploration}}} &
\multicolumn{2}{c}{ \textbf{Regret bound of}\, $\AlgName(\Alg, \Tcal_e)$ } \\
    \cmidrule(lr){2-3}
    & $\Alg$ = \texttt{LinUCB} & $\Alg$ = \texttt{LinTS} \\
    \midrule
    $f(t) = t$\, (\textit{base})
    & $\Ocal \left( \frac{1}{\Delta} \left( \log T + d \log \log T \right)^2 \right)$
    & $\Ocal \left( \frac{1}{\Delta}  \left( d \log T \right)\left( \log T + d \log \log T \right)^2 \right)$
    \\
    \midrule
    $f(t) = t / m$
    & $\Ocal\left( \left( m + \frac{d}{\Delta} \right) \frac{d}{\Delta} \log^2 \frac{d}{\Delta}\right)$
    & $\Ocal\left( \frac{d^3}{\Delta^2} \log^3 \frac{d}{\Delta} + \frac{md}{\Delta} \log^2 \frac{d}{\Delta} \right)$
    \\
    \midrule
    $f(t) = t / (\log t)^r$
    & $\Ocal \left( \frac{d^2}{\Delta^2}\log^{2+r} \frac{d}{\Delta} \right)$
    & $\Ocal\left( \frac{d^3}{\Delta^2}\log^{3+r} \frac{d}{\Delta} \right)$
    \\
    \midrule
    $f(t) = t^r$
    & $\Ocal \left( \left(\frac{d}{\Delta}\log \frac{d}{\Delta}\right)^{\frac{2}{r}} \right)$
    & $\Ocal \left( \left(\frac{d^{3}}{\Delta^{2}}\log^{3} \frac{d}{\Delta}\right)^{\frac{1}{r}} \right)$
    \\
    \midrule
    $f(t) = (\log t)^r$ & $ e^{ \Ocal \left( \left( \frac{d}{\Delta} \log \frac{d}{\Delta} \right)^{\frac{2}{r}} + \left( \frac{d}{\Delta^2} \right)^{\frac{1}{r-1}} \right)}$ & $e^{ \Ocal\left( \left( \frac{d^3}{\Delta^2}\log ^3\frac{d}{\Delta} \right)^{\frac{1}{r}} + \left( \frac{d}{\Delta^2} \right)^{\frac{1}{r-1}} \right)}$
    \\
    \bottomrule
\end{tabular}
\label{table:total regret}
\end{center}
\end{table}

\paragraph{Computational complexity.}
The computational time complexity of a single greedy selection is $\Ocal(d^2 + dK)$:
using the Sherman-Morrison formula~\citep{sherman1950adjustment}, one can maintain $V_t^{-1}$ in $\Ocal(d^2)$ time per step, so updating $\hat{\theta}_t$ also takes $\Ocal(d^2)$ time, and the remaining $\Ocal(dK)$ is required to find the arm with the highest estimated reward.
The computational complexity of $\texttt{LinUCB}$ is $\Ocal(d^2 + d^2K)$ per time step, where the additional $\Ocal(d^2K)$ term is required to compute the upper confidence bound of rewards $x^\top \hat{\theta}_t + \beta_t \| x \|_{V_t^{-1}}$ for all $x \in \Xcal$.
The computational complexity of $\texttt{LinTS}$ is $\Ocal(d^3 + dK)$, where the additional $\Ocal(d^3)$ term corresponds to sampling parameter $\tilde{\theta}_t$ from a multivariate Gaussian distribution.
Both algorithms have strictly greater computational complexity than performing a greedy selection, meaning that replacing them with greedy selections reduces the total computational cost.

\subsection{Necessity of \texorpdfstring{$\omega(\log t)$}{w(log t)} Exploration.}
We provide a lower-bound result that implies the condition $f(t) = \omega(\log t)$ is necessary to obtain a polylogarithmic regret bound that holds for any $T$.
Specifically, we show that if $f(t) = \omega(\log t)$ does not hold, that is, either the limit $\lim_{t \rightarrow \infty} \frac{\log t}{f(t)}$ does not exist or is above zero, then there exists a problem instance such that the regret of $\AlgName$ scales almost linearly in $T$ using the standard information-theoretical method.

\begin{theorem}
\label{thm:lower bound}
    Let $\Alg$ be an arbitrary policy and $\Tcal_e \subset \NN$ be a set of natural numbers.
    If $f(t) \ne \omega(\log t)$, then for an arbitrary constant $\varepsilon \in (0, 1)$, there exists a problem instance $(\Xcal, \theta^*)$ and a constant $c(f, \varepsilon) > 0$ that depends on $f$ and $\varepsilon$ such that 
    \begin{align*}
        \EE \left[ \Rcal_{\Greed{\Alg, \Tcal_e }}(T) \right] \ge c(f, \varepsilon) T^{1 - \varepsilon}
    \end{align*}
    for infinitely many $T \in \NN$.
\end{theorem}

We note that this result applies to predetermined exploration schedules, and the $\omega(\log t)$ threshold might not be necessary when the exploration schedule is adaptive to the observations.

The proof of \cref{thm:lower bound} is presented in \cref{appx:lower bound}.

\subsection{Sketch of Proof}
\label{sec:sketch of proof}

In this subsection, we provide a sketch of the proof of \cref{thm:main theorem}.
Throughout this subsection, we work under the high-probability event that $\Rcal_{\Alg}(T)$ is polylogarithmic in $T$ and the event of \cref{lma:self normalization} that ensures the concentration of $\hat{\theta}_t$ toward $\theta^*$.

We first explain how $\tau_{\Alg}$ is chosen.
Assuming that $\Alg$ is independently run, $\tau_{\Alg}$ is defined as the time step such that for all $T \ge \tau_{\Alg}$, at least a quarter of the selections made by $\Alg$ are optimal, that is, the optimal arm is chosen in at least $T / 4$ of the $T$ time steps.
The existence and order of $\tau_{\Alg}$ are guaranteed by the following lemma:

\begin{lemma}
\label{lma:large enough t}
    Suppose a linear bandit algorithm $\Alg'$ attains a polylogarithmic regret bound of $\Rcal_{\Alg'}(T) = \Ocal \left( \frac{d^a}{\Delta^b} \log^c T \right)$ for some constants $a, b, c \ge 0$.
    Then, there exists $\tau_{\Alg'} \in \NN$ such that for all $T \ge \tau_{\Alg'}$, at least a quarter of the $T$ selections made by $\Alg'$ are optimal.
    Furthermore, $\tau_{\Alg'} = \Ocal \left( \frac{d^a}{\Delta^{b+1}} \log^c \frac{d}{\Delta} \right)$.
\end{lemma}

We mainly focus on the sum of regret incurred after the time step $f^{-1}(\tau_{\Alg})$, that 
 is, after $\Alg$ is executed for $\tau_{\Alg}$ times.
For $\tau, T \in \NN$, let $\Gcal(\tau, T) := \{ t \in \NN : \tau + 1 \le t \le T, t \notin \Tcal_e \}$, which denotes the set of time steps with greedy selections between $\tau + 1$ and $T$, inclusively.
Let $\Rcal_{\Greed{\Alg, \Tcal_e}}^G(\tau, T) := \sum_{t \in \Gcal(\tau, T)} \text{reg}_t$ be the cumulative regret incurred at the time steps in $\Gcal(\tau, T)$.
In the remainder of this section, we show that $\Rcal_{\Greed{\Alg, \Tcal_e}}^G(f^{-1}(\tau_{\Alg}) + \tau_1, T)$ has the polylogarithmic bound stated in \cref{thm:main theorem} for some constant $\tau_1$.

The following lemma shows that the regret of greedy selections is related to the number of optimal selections.

\begin{lemma}
\label{lma:first decomposition}
    For any $\tau, T \in \NN$ with $\tau < T$, it holds that
    \begin{align*}
        \Rcal_{\Greed{\Alg, \Tcal_e}}^G (\tau, T) \le \frac{4 \alpha_T \beta_T^2}{\Delta} + \frac{2}{\Delta} \sum_{t \in \Gcal(\tau, T)} \frac{\beta_{t-1}^2}{1 + \Nopt(t - 1)}
        \, .
    \end{align*}
\end{lemma}

The intuition behind this lemma is that the estimator $\hat{\theta}_t$ becomes more accurate in estimating $x^{*\top}\theta^*$ as the optimal arm $x^*$ is selected more often.
The conclusion of the lemma implies that if $\Nopt(t)$ increases linearly in $t$, then the additional regret caused by the greedy selections remains polylogarithmic in $T$.
By the choice of $\tau_{\Alg}$, at least a quarter of the selections made by $\Alg$ are optimal for all $t \ge f^{-1}(\tau_{\Alg})$, implying that $\Nopt(t) \ge \frac{1}{4} f(t)$.
This fact leads to the following regret bound:

\begin{lemma}
\label{prop:weak regret bound}
    Let $\tau_{\Alg}$ be defined as in \cref{thm:main theorem}.
    Then, for any $T > f^{-1}(\tau_{\Alg})$, it holds that
    \begin{align*}
        \Rcal_{\Greed{\Alg, \Tcal_e}}^G(f^{-1}(\tau_{\Alg}), T) \le \frac{4 \alpha_T \beta_T^2}{\Delta} + \frac{8}{\Delta} \sum_{t \in \Gcal(f^{-1}(\tau_{\Alg}), T)} \frac{ \beta_t^2}{f(t)}
        \, .
    \end{align*}
    Furthermore, this bound is sublinear in $T$ when $f(t) = \omega( \log t)$.
\end{lemma}

We further improve this bound by observing that the quantity $\Nopt(t)$ must grow linearly with $t$ for sufficiently large $t$ as we now have a sublinear bound on $\Rcal_{\Greed{\Alg, \Tcal_e}}$.
Using this fact, we obtain the following stronger regret bound.

\begin{proposition}
\label{prop:strong regret bound}
    There exists a constant $\tau_1 \in \NN$ that depends on $d$, $\Delta$, $\tau_{\Alg}$, and the function $f$, is independent of $T$, and satisfies
    \begin{align*}
        \Rcal_{\Greed{\Alg, \Tcal_e}}^G(f^{-1}(\tau_{\Alg}), f^{-1}(\tau_{\Alg}) + \tau_1) \le \frac{7}{16} \Delta \tau_1
    \end{align*}
    and
    \begin{align*}
        \Rcal_{\Greed{\Alg, \Tcal_e}}^G(f^{-1}(\tau_{\Alg}) + \tau_1, T) \le \frac{4 \alpha_T \beta_T^2}{\Delta} + \frac{16 \beta_T^2 \log T}{\Delta}
    \end{align*}
    for all $T > f^{-1}(\tau_{\Alg}) + \tau_1$.
\end{proposition}

Note that $\beta_T^2 = \Ocal(\alpha_T)$, so we have derived a bound of $\Ocal( \alpha_T (\alpha_T + \log T) / \Delta)$ with some additional constant amount.
The proof is completed by providing an appropriate bound on $\alpha_T$.
We apply the following lemma, which is derived from the proof of Theorem 5 in~\citet{abbasi2011improved}.

\begin{lemma}
\label{lma:better bound on alpha}
    If the data $X_1, X_2, \ldots, X_T$ is collected through a linear bandit algorithm $\Alg'$, then
    \begin{align*}
        \alpha_T \le \log \left( 1 + T \right) + (d - 1) \log \left( 1 + \frac{\Rcal_{\Alg'}(T) }{(d - 1) \Delta} \right)
        \, .
    \end{align*}
    Consequently, if $\Alg'$ attains polylogarithmic regret, then
    \begin{align*}
        \alpha_T = \Ocal \left( \log T + d \log \log T  + d \log \frac{d}{\Delta} \right)
        \, .
    \end{align*}
\end{lemma}

The detailed proof of \cref{thm:main theorem} is presented in \cref{appx:proof main}.

\begin{remark}[]
    The analysis of Theorem 1 requires positivity of the minimum gap $\Delta$ and a fixed optimal arm.
    Therefore, the analysis holds as long as the two conditions are satisfied, even for infinite and time-varying arm sets, although it does not fully generalize to the linear contextual bandit setting with arbitrary arm sets.
    For a detailed discussion on the possibility of extending the analysis to time-varying arm sets, refer to \cref{appx:context}.
\end{remark}

\section{Numerical Experiments}
\label{sec:experiment}

\begin{figure}[!ht]
    \centering
    \includegraphics[width=\linewidth]{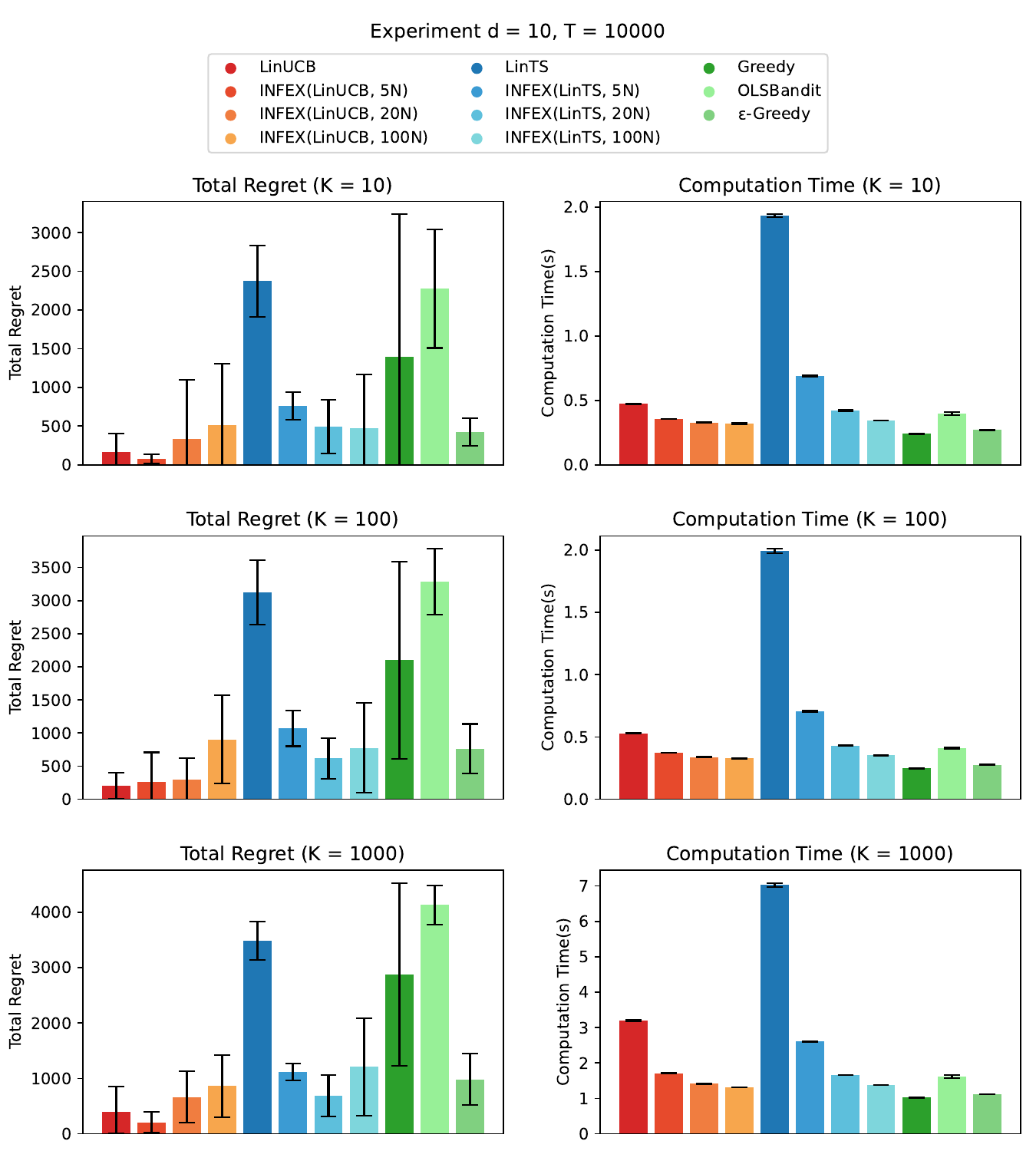}
    \caption{Comparison of total regret (left) and computation time (right) when $d = 10$, $T = 10000$, and $K = 10$ (top), $K = 100$ (middle), and $K = 1000$ (bottom).
    }
    \label{fig:d10}
\end{figure}

To complement our theoretical analysis, we conduct numerical simulations to empirically investigate the behavior and practical benefits of $\AlgName$. Our main objectives are to (i) assess whether infrequent exploration strategies maintain strong regret performance compared to fully adaptive methods, (ii) evaluate computational efficiency improvements due to reduced exploration frequency, and (iii) demonstrate the general applicability and robustness of our proposed framework across different base exploratory algorithms and exploration schedules.

We select $\Alg = \texttt{LinUCB}$ and $\Alg = \texttt{LinTS}$ as the base algorithms for exploration and use an exploration schedule $\Tcal_e = m \NN := \{ m n : n \in \NN\}$, meaning $\Alg$ executes every $m$ steps. Specifically, we examine three choices of $m$: $m = 5$, $m = 20$, and $m = 100$, corresponding to $80\%$, $95\%$, and $99\%$ greedy selections, respectively. 
For benchmarking, we also compare our framework against other policies: the purely greedy policy, a single-parameter version of $\texttt{OLSBandit}$~\citep{goldenshluger2013linear}, and an $\varepsilon$-$\texttt{greedy}$ approach with $\varepsilon_t = t^{-1/3}$.

We randomly generate problem instances for given $d$ and $K$ as follows.
We construct the arm set $\Xcal$ by sampling $K$ arms i.i.d. from a multivariate Gaussian distribution $\Ncal(\boldsymbol{0}_d, \frac{1}{2d} I_d)$ and rescaling each vector to have a norm at most 1 when it exceeds 1.
We sample $\theta^*$ uniformly from the unit sphere in $\RR^d$.
The random reward is given as either $+1$ or $-1$, with its expectation being $X_t^\top \theta^*$.
We repeat the process for 20 randomly generated instances and report the mean and standard deviation of the cumulative regret over $T = 10000$ time steps for each algorithm.

Figure~\ref{fig:d10} shows the total regret and computation time of each algorithm.
Interestingly, we observe that certain exploration schedules \textit{improve} the total regret.
Especially for $\Alg = \texttt{LinTS}$, all values of $m = 5, 20, 100$ reduce the regret significantly.
The performance of $\Alg = \texttt{LinUCB}$ is also improved when $m = 5$.
These configurations outperform both the base algorithm and the purely greedy policy, exhibiting strong practicality.
We also observe a reduction of computational time for any value of $m$.

$\texttt{OLSBandit}$ is inefficient because it spends most of the time steps, specifically at least $\Omega(d^2 \log T)$ steps, on forced sampling.
While $\varepsilon$-$\texttt{greedy}$ appears to show decent performance, we note that the choice $\varepsilon_t = t^{-1/3}$ implies a regret lower bound of $\Omega(T^{2/3})$ and it is its best bound, precluding the possibility of achieving polylogarithmic regret.

Refer to \cref{appx:experiment} for additional experiments with different dimensions $d$ and experiment details.

\section{Conclusion}
\label{sec:conclusion}

We propose $\AlgName$, a simple yet practical framework that mainly performs greedy selections while exploring according to a given schedule.
Our theoretical analysis reveals that $\AlgName$ attains a polylogarithmic regret bound, whose growth rate with respect to $T$ remains independent of the exploration schedule, provided that the exploration frequency exceeds the order of $\log T$.
Empirical results further illustrate the strengths of $\AlgName$, showing that judiciously timed exploration not only maintains robust theoretical performance guarantees but also delivers practical improvements in terms of both regret and computational efficiency.
While this work focuses specifically on linear bandit settings, we believe the framework and results serve as a foundation for broader exploration strategies, potentially enabling similar performance benefits in more complex and general function approximation scenarios. 
An exciting avenue for future research lies in extending our framework to accommodate these generalizations, further enhancing its applicability and impact.

\newpage

\section*{Acknowledgements}
This work was supported by the National Research Foundation of Korea~(NRF) grant funded by the Korea government~(MSIT) (No.  RS-2022-NR071853 and RS-2023-00222663), by Institute of Information \& communications Technology Planning \& Evaluation~(IITP) grant funded by the Korea government~(MSIT) (No. RS-2025-02263754), and by AI-Bio Research Grant through Seoul National University.

\bibliography{references}

\begin{thebibliography}{32}
\providecommand{\natexlab}[1]{#1}
\providecommand{\url}[1]{\texttt{#1}}
\expandafter\ifx\csname urlstyle\endcsname\relax
  \providecommand{\doi}[1]{doi: #1}\else
  \providecommand{\doi}{doi: \begingroup \urlstyle{rm}\Url}\fi

\bibitem[Abbasi-Yadkori et~al.(2009)Abbasi-Yadkori, Antos, and Szepesv{\'a}ri]{abbasi2009forced}
Yasin Abbasi-Yadkori, Andr{\'a}s Antos, and Csaba Szepesv{\'a}ri.
\newblock Forced-exploration based algorithms for playing in stochastic linear bandits.
\newblock In \emph{COLT Workshop on On-line Learning with Limited Feedback}, volume~92, page 236, 2009.

\bibitem[Abbasi-Yadkori et~al.(2011)Abbasi-Yadkori, P{\'a}l, and Szepesv{\'a}ri]{abbasi2011improved}
Yasin Abbasi-Yadkori, D{\'a}vid P{\'a}l, and Csaba Szepesv{\'a}ri.
\newblock Improved algorithms for linear stochastic bandits.
\newblock \emph{Advances in Neural Information Processing Systems}, 24:\penalty0 2312--2320, 2011.

\bibitem[Abe and Long(1999)]{abe1999associative}
Naoki Abe and Philip~M Long.
\newblock Associative reinforcement learning using linear probabilistic concepts.
\newblock In \emph{International Conference on Machine Learning}, pages 3--11, 1999.

\bibitem[Abeille and Lazaric(2017)]{abeille2017Linear}
Marc Abeille and Alessandro Lazaric.
\newblock {Linear Thompson Sampling Revisited}.
\newblock In \emph{Proceedings of the 20th International Conference on Artificial Intelligence and Statistics}, volume~54, pages 176--184. PMLR, PMLR, 2017.

\bibitem[Agrawal and Goyal(2012)]{agrawal2012analysis}
Shipra Agrawal and Navin Goyal.
\newblock Analysis of thompson sampling for the multi-armed bandit problem.
\newblock In \emph{Conference on learning theory}, pages 39--1. JMLR Workshop and Conference Proceedings, 2012.

\bibitem[Agrawal and Goyal(2013)]{agrawal2013thompson}
Shipra Agrawal and Navin Goyal.
\newblock Thompson sampling for contextual bandits with linear payoffs.
\newblock In \emph{International conference on machine learning}, pages 127--135. PMLR, 2013.

\bibitem[Auer(2002)]{auer2002using}
Peter Auer.
\newblock Using confidence bounds for exploitation-exploration trade-offs.
\newblock \emph{Journal of Machine Learning Research}, 3\penalty0 (Nov):\penalty0 397--422, 2002.

\bibitem[Auer et~al.(2002)Auer, Cesa-Bianchi, and Fischer]{auer2002finite}
Peter Auer, Nicolo Cesa-Bianchi, and Paul Fischer.
\newblock Finite-time analysis of the multiarmed bandit problem.
\newblock \emph{Machine learning}, 47\penalty0 (2):\penalty0 235--256, 2002.

\bibitem[Bastani and Bayati(2020)]{bastani2020online}
Hamsa Bastani and Mohsen Bayati.
\newblock Online decision making with high-dimensional covariates.
\newblock \emph{Operations Research}, 68\penalty0 (1):\penalty0 276--294, 2020.

\bibitem[Bastani et~al.(2021)Bastani, Bayati, and Khosravi]{bastani2021mostly}
Hamsa Bastani, Mohsen Bayati, and Khashayar Khosravi.
\newblock Mostly exploration-free algorithms for contextual bandits.
\newblock \emph{Management Science}, 67\penalty0 (3):\penalty0 1329--1349, 2021.

\bibitem[Bretagnolle and Huber(1979)]{bretagnolle1979estimation}
Jean Bretagnolle and Catherine Huber.
\newblock Estimation des densit{\'e}s: risque minimax.
\newblock \emph{Zeitschrift f{\"u}r Wahrscheinlichkeitstheorie und verwandte Gebiete}, 47:\penalty0 119--137, 1979.

\bibitem[Dani et~al.(2008)Dani, Hayes, and Kakade]{dani2008stochastic}
Varsha Dani, Thomas~P Hayes, and Sham~M Kakade.
\newblock Stochastic linear optimization under bandit feedback.
\newblock In \emph{21st Annual Conference on Learning Theory}, number 101, pages 355--366, 2008.

\bibitem[Garivier et~al.(2016)Garivier, Lattimore, and Kaufmann]{garivier2016explore}
Aur{\'e}lien Garivier, Tor Lattimore, and Emilie Kaufmann.
\newblock On explore-then-commit strategies.
\newblock \emph{Advances in Neural Information Processing Systems}, 29, 2016.

\bibitem[Goldenshluger and Zeevi(2013)]{goldenshluger2013linear}
Alexander Goldenshluger and Assaf Zeevi.
\newblock A linear response bandit problem.
\newblock \emph{Stochastic Systems}, 3\penalty0 (1):\penalty0 230--261, 2013.

\bibitem[Hanna et~al.(2023)Hanna, Yang, and Fragouli]{hanna2023contexts}
Osama~A Hanna, Lin Yang, and Christina Fragouli.
\newblock Contexts can be cheap: Solving stochastic contextual bandits with linear bandit algorithms.
\newblock In \emph{The Thirty Sixth Annual Conference on Learning Theory}, pages 1791--1821. PMLR, 2023.

\bibitem[Hao et~al.(2020)Hao, Lattimore, and Wang]{hao2020high}
Botao Hao, Tor Lattimore, and Mengdi Wang.
\newblock High-dimensional sparse linear bandits.
\newblock \emph{Advances in Neural Information Processing Systems}, 33:\penalty0 10753--10763, 2020.

\bibitem[Hsu et~al.(2012)Hsu, Kakade, and Zhang]{hsu2012random}
Daniel Hsu, Sham~M Kakade, and Tong Zhang.
\newblock Random design analysis of ridge regression.
\newblock In \emph{Conference on learning theory}, pages 9--1. JMLR Workshop and Conference Proceedings, 2012.

\bibitem[Jedor et~al.(2021)Jedor, Lou{\"e}dec, and Perchet]{jedor2021greedy}
Matthieu Jedor, Jonathan Lou{\"e}dec, and Vianney Perchet.
\newblock Be greedy in multi-armed bandits.
\newblock \emph{arXiv preprint arXiv:2101.01086}, 2021.

\bibitem[Jin et~al.(2023)Jin, Yang, Xiao, and Xu]{jin2023thompson}
Tianyuan Jin, Xianglin Yang, Xiaokui Xiao, and Pan Xu.
\newblock Thompson sampling with less exploration is fast and optimal.
\newblock In \emph{International Conference on Machine Learning}, pages 15239--15261. PMLR, 2023.

\bibitem[Kannan et~al.(2018)Kannan, Morgenstern, Roth, Waggoner, and Wu]{kannan2018smoothed}
Sampath Kannan, Jamie~H Morgenstern, Aaron Roth, Bo~Waggoner, and Zhiwei~Steven Wu.
\newblock A smoothed analysis of the greedy algorithm for the linear contextual bandit problem.
\newblock \emph{Advances in neural information processing systems}, 31, 2018.

\bibitem[Kim and Oh(2024)]{kim2024local}
Seok-Jin Kim and Min-hwan Oh.
\newblock Local anti-concentration class: Logarithmic regret for greedy linear contextual bandit.
\newblock \emph{Advances in Neural Information Processing Systems}, 37:\penalty0 77525--77592, 2024.

\bibitem[Langford and Zhang(2007)]{langford2007epoch}
John Langford and Tong Zhang.
\newblock The epoch-greedy algorithm for multi-armed bandits with side information.
\newblock \emph{Advances in neural information processing systems}, 20, 2007.

\bibitem[Lattimore and Szepesv{\'a}ri(2020)]{lattimore2020bandit}
Tor Lattimore and Csaba Szepesv{\'a}ri.
\newblock \emph{Bandit algorithms}.
\newblock Cambridge University Press, 2020.

\bibitem[Lee et~al.(2025)Lee, Hwang, and hwan Oh]{lee2025lasso}
Harin Lee, Taehyun Hwang, and Min hwan Oh.
\newblock Lasso bandit with compatibility condition on optimal arm.
\newblock In \emph{The Thirteenth International Conference on Learning Representations}, 2025.

\bibitem[Oh et~al.(2021)Oh, Iyengar, and Zeevi]{oh2021sparsity}
Min-hwan Oh, Garud Iyengar, and Assaf Zeevi.
\newblock Sparsity-agnostic lasso bandit.
\newblock In \emph{International Conference on Machine Learning}, pages 8271--8280. PMLR, 2021.

\bibitem[Perchet et~al.(2016)Perchet, Rigollet, Chassang, and Snowberg]{perchet2016batched}
Vianney Perchet, Philippe Rigollet, Sylvain Chassang, and Erik Snowberg.
\newblock Batched bandit problems.
\newblock 2016.

\bibitem[Raghavan et~al.(2023)Raghavan, Slivkins, Vaughan, and Wu]{raghavan2023greedy}
Manish Raghavan, Aleksandrs Slivkins, Jennifer~Wortman Vaughan, and Zhiwei~Steven Wu.
\newblock Greedy algorithm almost dominates in smoothed contextual bandits.
\newblock \emph{SIAM Journal on Computing}, 52\penalty0 (2):\penalty0 487--524, 2023.

\bibitem[Sherman and Morrison(1950)]{sherman1950adjustment}
Jack Sherman and Winifred~J Morrison.
\newblock Adjustment of an inverse matrix corresponding to a change in one element of a given matrix.
\newblock \emph{The Annals of Mathematical Statistics}, 21\penalty0 (1):\penalty0 124--127, 1950.

\bibitem[Sivakumar et~al.(2020)Sivakumar, Wu, and Banerjee]{sivakumar2020structured}
Vidyashankar Sivakumar, Steven Wu, and Arindam Banerjee.
\newblock Structured linear contextual bandits: A sharp and geometric smoothed analysis.
\newblock In \emph{International Conference on Machine Learning}, pages 9026--9035. PMLR, 2020.

\bibitem[Thompson(1933)]{thompson1933likelihood}
William~R Thompson.
\newblock On the likelihood that one unknown probability exceeds another in view of the evidence of two samples.
\newblock \emph{Biometrika}, 25\penalty0 (3/4):\penalty0 285--294, 1933.

\bibitem[Tirinzoni et~al.(2022)Tirinzoni, Papini, Touati, Lazaric, and Pirotta]{tirinzoni2022scalable}
Andrea Tirinzoni, Matteo Papini, Ahmed Touati, Alessandro Lazaric, and Matteo Pirotta.
\newblock Scalable representation learning in linear contextual bandits with constant regret guarantees.
\newblock \emph{Advances in Neural Information Processing Systems}, 35:\penalty0 2307--2319, 2022.

\bibitem[Weyl(1912)]{weyl1912asymptotische}
Hermann Weyl.
\newblock Das asymptotische verteilungsgesetz der eigenwerte linearer partieller differentialgleichungen (mit einer anwendung auf die theorie der hohlraumstrahlung).
\newblock \emph{Mathematische Annalen}, 71\penalty0 (4):\penalty0 441--479, 1912.

\end{thebibliography}
\bibliographystyle{plainnat}

\newpage
\appendix

\section{Proof of Theorem~\ref{thm:main theorem}}
\label{appx:proof main}

In this section, we provide a detailed proof of \cref{thm:main theorem}.
We supplement the proof by proving \cref{prop:strong regret bound} in \cref{appx:proof of prop} and verifying the bounds of $\constterm{\tau_{\Alg}, f}$ listed in \cref{table:const term} in \cref{appx:constant term}.
In \cref{appx:lower bound}, we prove \cref{thm:lower bound}.
Proofs of technical lemmas are provided in \cref{appx:proof lemma}.

Throughout the proof, we denote $\tau_0 := f^{-1}(\tau_{\Alg})$ for simplicity.

\subsection{Proof of Theorem~\ref{thm:main theorem}}

\begin{proof}[Proof of Theorem~\ref{thm:main theorem}]
$\tau_{\Alg}$ is set in the way described in \cref{lma:large enough t} with $\Alg' = \Alg$, and the lemma guarantees that $\tau_{\Alg} = \Ocal( \frac{d^a}{\Delta^{b+1}} \log^c \frac{d}{\Delta})$.
$\tau_1$ is the constant defined in \cref{prop:strong regret bound}.

The total regret is decomposed into four parts, described in Eq.~\eqref{eq:regret decomposition}.

\begin{align}
\label{eq:regret decomposition}
    \Rcal_{\Greed{\Alg, \Tcal_e}}(T) & \le \Rcal_{\Alg}(f(T)) + 2 S \tau_0 + \Rcal_{\Greed{\Alg, \Tcal_e}}^G(\tau_0, \tau_0 + \tau_1 )
    \nonumber
    \\
    & \qquad
    + \Rcal_{\Greed{\Alg, \Tcal_e}}^G(\tau_0 + \tau_1, T )
    \, .
\end{align}

The first term is the sum of the regret incurred by $\Alg$.
Since $\Alg$ is executed $f(T)$ times, this regret is bounded by $\Rcal_{\Alg}(f(T))$.
The second part is the sum of the regret incurred by the greedy selections during the first $\tau_0$ time steps.
Since the maximum possible regret per time step is $2S$, we bound the sum by $2S \tau_0$.
Note that this quantity is independent of $T$.
Lastly, among the time steps that perform greedy selections, $\Rcal_{\Greed{\Alg, \Tcal_e}}^G(\tau_0, \tau_0 + \tau_1)$ is the sum of the regret incurred during the time steps between $\tau_0 + 1$ and $\tau_0 + \tau_1$, inclusively, and $\Rcal_{\Greed{\Alg, \Tcal_e}}^G(\tau_0 + \tau_1, T)$ is the sum of the regret incurred during the time steps between $\tau_0 + \tau_1 + 1$ and $T$, inclusively.
\\
By \cref{prop:strong regret bound}, we have $\Rcal_{\Greed{\Alg, \Tcal_e}}^G(\tau_0, \tau_0 + \tau_1 ) \le \frac{7}{16} \Delta \tau_1$ and $\Rcal_{\Greed{\Alg, \Tcal_e}}^G(\tau_0 + \tau_1, T ) = \Ocal(\alpha_T (\alpha_T + \log T) / \Delta)$.
Denoting $\tilde{G}_{\text{const}} := 2S \tau_0 + \frac{7}{16} \Delta \tau_1$, we obtain that
\begin{align}
    \Rcal_{\Greed{\Alg, \Tcal_e}}(T) & \le \Rcal_{\Alg}(f(T))
    + \tilde{G}_{\text{const}} + \Ocal \left( \frac{\alpha_T ( \alpha_T + \log T)}{\Delta} \right)
    \label{eq:regret bound 1}
    \\
    & = \Rcal_{\Alg}(f(T))
    + \tilde{G}_{\text{const}} + \Ocal \left( \frac{(d \log T)^2}{\Delta} \right)
    \label{eq:regret bound 2}
    \, ,
\end{align}
where we use \cref{lma:bound on alpha} for the last equality.
Eq.~\eqref{eq:regret bound 2} shows that $\Greed{\Alg, \Tcal_e}$ achieves a polylogarithmic regret bound added by a $T$-independent constant.
We improve the bound on $\alpha_T$ using \cref{lma:better bound on alpha} and the derived regret bound.
The growth rate of the logarithm of the cumulative regret is $\log (1 + \Rcal_{\Greed{\Alg, \Tcal_e}}(T)) = \Ocal( \log (\frac{d}{\Delta} \log T) + \log \tilde{G}_{\text{const}})$.
Applying this fact to \cref{lma:better bound on alpha}, we obtain that
\begin{align*}
    \alpha_T = \Ocal \left( \log T + d \log \log T + d \log\frac{d}{\Delta} + d\log \tilde{G}_{\text{const}} \right)
    \, .
\end{align*}
Plugging this bound into Eq.~\eqref{eq:regret bound 1}, we obtain that
\begin{align}
    \Rcal_{\Greed{\Alg, \Tcal_e}}(T) & \le \Rcal_{\Alg}(f(T))
    + \tilde{G}_{\text{const}}
    \nonumber
    \\
    & \qquad + \Ocal \left( \frac{1}{\Delta} \left( \log T + d \log \log T + d \log\frac{d}{\Delta} + d\log \tilde{G}_{\text{const}} \right)^2\right)
    \nonumber
    \\
    & = \Rcal_{\Alg}(f(T))
    + \tilde{G}_{\text{const}} + \Ocal\left( \frac{1}{\Delta}  \left( d\log \tilde{G}_{\text{const}} \right)^2 \right)
    \nonumber
    \\
    & \qquad + \Ocal \left( \frac{1}{\Delta} \left( \log T + d \log \log T + d \log\frac{d}{\Delta} \right)^2\right)
    \label{eq:regret bound 3}
    \, ,
\end{align}
where the last equality holds since $(a+b)^2 \le 2 a^2 + 2b^2$ for all $a, b \in \RR$.
Therefore, there exists a constant $\constterm{\tau_{\Alg}, f} =\tilde{G}_{\text{const}} + \Ocal\left( \frac{1}{\Delta} \left( d\log \tilde{G}_{\text{const}} \right)^2 \right)$ and a function $G(T)$ in $\Ocal \left( \frac{1}{\Delta} \left( \log T + d \log \log T + d \log\frac{d}{\Delta} \right)^2\right)$ such that 
\begin{align*}
    \Rcal_{\Greed{\Alg, \Tcal_e}}(T) \le \Rcal_{\Alg}(f(T)) + \constterm{\tau_{\Alg}, f} + G(T)
    \, .
\end{align*}
In \cref{appx:constant term}, we summarize how $\constterm{\tau_{\Alg}, f}$ is determined and provide its example bounds listed in \cref{table:const term}.
\end{proof}

\subsection{Proof of Proposition~\ref{prop:strong regret bound}}
\label{appx:proof of prop}

\begin{proof}[Proof of Proposition~\ref{prop:strong regret bound} ]
    By the sublinearity stated in \cref{prop:weak regret bound},
    there exists a constant $\tau_1$ that depends on $d, \Delta, \tau_{\Alg}$, and $f$ such that for all $T \ge \tau_1$,
    \begin{align}
    \label{eq:definition of tau1}
        \frac{4 \alpha_T \beta_T^2}{\Delta} + \frac{8}{\Delta} \sum_{t \in \Gcal(\tau_0, T)} \frac{\beta_t^2}{f(t)}
        \le \frac{7}{16} \Delta (T - \tau_0)
        \, ,
    \end{align}
    The first part of the proposition is trivial by the choice of $\tau_1$.
    Now, we prove the second part.
    Fix $T > \tau_0 + \tau_1$.
    While \cref{prop:weak regret bound} only considers the optimal selections by $\Alg$, we improve this result by showing that the number of optimal selections grows linearly in $T$ and combining it with \cref{lma:first decomposition}.
    Specifically, we show that $\Nopt(T) \ge \frac{1}{8}(T - \tau_0)$.
    We consider two cases.
    First, suppose $\Alg$ is executed at more than half of the time steps between $\tau_0 + 1$ and $T$, that is, $|\Tcal_e \cap \{ \tau_0 + 1, \ldots, T\} | \ge \frac{1}{2} (T - \tau_0)$.
    Then, $f(T) \ge \frac{1}{2} (T - \tau_0)$.
    Since at least a quarter of the selections made by $\Alg$ are optimal after time step $t = \tau_0$, it holds that
    \begin{align*}
        \Nopt(T) 
        & \ge \frac{1}{4} f(T) \ge \frac{1}{8} (T - \tau_0)
        \, .
    \end{align*}
    Now, we suppose the opposite. 
    Consider the case where $\Alg$ is executed at fewer than half of the time steps between $t = \tau_0 + 1$ and $T$.
    Then, $\frac{1}{2}(T - \tau_0) \le | \Gcal(\tau_0, T) |$.
    We bound the number of suboptimal selections during the time steps in $\Gcal(\tau_0, T)$ as follows:
    \begin{align*}
        \sum_{t \in \Gcal(\tau_0, T)} \Delta \ind\{ X_t \ne x^* \}
        & \le \Rcal_{\Greed{\Alg, \Tcal_e}}^G(\tau_0, T)
        \\
        & \le \frac{4 \alpha_T \beta_T^2}{\Delta} + \frac{8}{\Delta} \sum_{t \in \Gcal(\tau_0, T)} \frac{\beta_t^2}{f(t)}
        \\
        & \le \frac{7}{16} \Delta (T - \tau_0)
        \\
        & \le \frac{7}{8} \Delta \left|  \Gcal(\tau_0, T)) \right|
        \, ,
    \end{align*}
    where the first inequality uses that the non-zero instantaneous regret is at least $\Delta$, the second inequality applies \cref{prop:weak regret bound}, the third inequality follows from Eq.~\eqref{eq:definition of tau1}, and the last inequality uses that $\frac{1}{2}(T - \tau_0) \le | \Gcal(\tau_0, T)|$.
    Therefore, we conclude that the number of suboptimal selections at time steps in $\Gcal(\tau_0, T)$ is at most $\frac{7}{8} | \Gcal(\tau_0, T) |$.
    It follows that the number of optimal selections among the same set of time steps is at least $\frac{1}{8} | \Gcal(\tau_0, T) |$.
    Since at least a quarter of the exploratory selections are optimal, we have 
    \begin{align*}
        \Nopt(T) & \ge \frac{1}{8}|  \Gcal(\tau_0, T) | + \frac{1}{4} f(T)
        \\
        & \ge \frac{1}{8}|  \Gcal(\tau_0, T) | + \frac{1}{8} (f(T) - \tau_{\Alg})
        \\
        & = \frac{1}{8} (T - \tau_0) 
        \, ,
    \end{align*}
    where the last equality comes from that $|\Gcal(\tau_0, T)|$ and $f(T) - \tau_{\Alg}$ are the numbers of greedy selections and exploratory selections during time steps $t = \tau_0 + 1, \ldots, T$ respectively and hence their sum is $T - \tau_0$.
    We have proved that $\Nopt(T) \ge \frac{1}{8} (T - \tau_0) $ for both cases.
    Plugging this bound into \cref{lma:first decomposition}, we conclude that
    \begin{align*}
        \frac{2}{\Delta} \sum_{t \in \Gcal(\tau_0 + \tau_1, T)} \frac{\beta_t^2}{1 + \Nopt(t - 1)}
        & \le \frac{2}{\Delta} \sum_{t \in \Gcal(\tau_0 + \tau_1, T)} \frac{\beta_t^2}{\frac{1}{8}(t - \tau_0)}
        \\
        & \le \frac{16 \beta_T^2}{\Delta} \sum_{t \in \Gcal(\tau_0 + \tau_1, T)} \frac{1}{t - \tau_0}
        \\
        & \le \frac{16 \beta_T^2}{\Delta} \int_{\tau_0 + \tau_1}^T \frac{1}{x - \tau_0} \, dx
        \\
        & = \frac{16 \beta_T^2 (\log (T - \tau_0) - \log \tau_1)}{\Delta}
        \\
        & \le \frac{16 \beta_T^2 \log T }{\Delta}
        \, ,
    \end{align*}
    where the first inequality holds since $1 + \Nopt(t - 1) \ge 1 + \frac{1}{8}(t - 1 - \tau_0) \ge \frac{1}{8}(t - \tau_0)$, the second inequality uses that $\beta_t$ is increasing, and the third inequality upper bounds the summation by an integral since $1/(t - \tau_0)$ is decreasing in $t$.
    The proof is completed by plugging this bound into \cref{lma:first decomposition}.
    \begin{align*}
        \Rcal_{\Greed{\Alg, \Tcal_e}}^G(\tau_0 + \tau_1, T) 
        & \le \frac{4\alpha_T \beta_T^2}{\Delta} + \frac{2}{\Delta} \sum_{t \in \Gcal(\tau_0 + \tau_1, T)} \frac{\beta_t^2}{1 + \Nopt(t - 1)}
        \\
        & = \frac{4\alpha_T \beta_T^2}{\Delta} + \frac{16 \beta_T^2 \log T }{\Delta}
        \, .
    \end{align*}
\end{proof}

\subsection{Bounds on \texorpdfstring{$\constterm{\tau_{\Alg}, f}$}{}}
\label{appx:constant term}

In this subsection, we provide bounds on $\constterm{\tau_{\Alg}, f}$.
The steps of determining $\constterm{\tau_{\Alg}, f}$ in the proofs of \cref{thm:main theorem} can be summarized as follows.
First, take $\tau_1$ such that for all $T \ge \tau_1$, it holds that
\begin{align*}
    \frac{4 \alpha_T \beta_T^2}{\Delta} + \frac{8}{\Delta} \sum_{t \in \Gcal(\tau_0, T)} \frac{\beta_t^2}{f(t)}
    \le \frac{7}{16} \Delta (T - \tau_0)
    \, ,
\end{align*}
which exists by \cref{prop:weak regret bound}.
Then, define $\tilde{G}_{\text{const}} := 2S \tau_0 + \frac{7}{16}\Delta \tau_1$.
Lastly, take $\constterm{\tau_{\Alg}, f} = \tilde{G}_{\text{const}} + \Ocal( \frac{1}{\Delta}(d \log \tilde{G}_{\text{const}})^2)$.
The value of $\tau_0 = f^{-1}(\tau_{\Alg})$ is determined once $f$ and $\tau_{\Alg}$ are determined.
It remains to provide an upper bound for $\tau_1$.
We define additional constants whose bounds are easier to obtain.
Let $\tau_{1, 1} \in \NN$ be the least time step such that $\tau_{1, 1} \ge \tau_0$ and for all $T \ge \tau_0 + \tau_{1, 1}$, it holds that
\begin{align*}
    \frac{4\alpha_T \beta_T^2}{\Delta} \le \frac{1}{16} \Delta T
    \, .
\end{align*}
Since $\alpha_T, \beta_T^2 = \Ocal(d \log T)$, we infer that $\tau_{1, 1} = \max\{ \tau_0, \Ocal( (\frac{d}{\Delta} \log \frac{d}{\Delta})^2) \} = \Ocal(\tau_0 + (\frac{d}{\Delta} \log \frac{d}{\Delta})^2)$.
Define $\tau_{1, 2} \in \NN$ to be the least time step such that for all $T \ge \tau_0 + \tau_{1, 2}$, it holds that
\begin{align*}
    \frac{8}{\Delta} \sum_{t \in \Gcal(\tau_0, T)} \frac{\beta_t^2}{f(t)} \le \frac{1}{4} \Delta (T - \tau_0)
    \, .
\end{align*}
The scale of $\tau_{1, 2}$ depends on $f(t)$.
Putting together, we obtain that for all $T \ge \tau_0 + \max\{ \tau_{1, 1}, \tau_{1, 2} \}$, it holds that
\begin{align*}
    \frac{4 \alpha_T \beta_T^2}{\Delta} + \frac{8}{\Delta} \sum_{t \in \Gcal(\tau_0, T)} \frac{\beta_t^2}{f(t)}
    & \le \frac{1}{16} \Delta T + \frac{1}{4} \Delta (T - \tau_0)
    \\
    & \le \frac{3}{8} \Delta (T - \tau_0)
    \\
    & \le \frac{7}{16} \Delta (T - \tau_0)
    \, ,
\end{align*}
where we use $T \le 2(T - \tau_0)$ for the second inequality, which is implied by $T \ge \tau_0 + \tau_{1, 1}\ge 2 \tau_0$.
Since $\tau_1$ is the least value that satisfies the property above, we have $\tau_1 \le \max\{ \tau_{1, 1}, \tau_{1, 2}\}$.
Then, we obtain that $\tilde{G}_{\text{const}} = \Ocal( \tau_0 + \Delta \tau_{1, 2} + \frac{d^2}{\Delta} \log^2 \frac{d}{\Delta}) $.
Additionally, note that for some universal constant $C> 0$, we have $\frac{d^2}{\Delta}\log^2 x \le x$ for all $x \ge \frac{C}{\Delta}\left(d\log\frac{d}{\Delta}\right)^2$.
Therefore, we have $\frac{d^2}{\Delta} \log^2 x = \Ocal ( x + \frac{d^2}{\Delta} \log^2 \frac{d}{\Delta})$.
It implies that
\begin{align*}
    \tilde{G}_{\text{const}} + \Ocal \left( \frac{1}{\Delta} (d \log \tilde{G}_{\text{const}})^2 \right) 
    & = \tilde{G}_{\text{const}} + \Ocal \left(\tilde{G}_{\text{const}} + \frac{d^2}{\Delta} \log^2 \frac{d}{\Delta}\right)
    \\
    & = \Ocal \left( \tau_0 + \Delta \tau_{1, 2} + \frac{d^2}{\Delta} \log^2 \frac{d}{\Delta} \right)
    \, .
\end{align*}
Combining with Eq.~\eqref{eq:regret bound 3} in the proof of \cref{thm:main theorem}, we obtain that
\begin{align*}
    \Rcal_{\Greed{\Alg, \Tcal_e}}(T) 
    & \le \Rcal_{\Alg}(f(T)) + \Ocal \left( \tau_0 + \Delta \tau_{1, 2} + \frac{d^2}{\Delta} \log^2 \frac{d}{\Delta} \right)
    \\
    & \qquad + \Ocal \left( \frac{1}{\Delta} \left( \log T + d \log \log T + d \log \frac{d}{\Delta} \right)^2 \right)
    \\
    & = \Rcal_{\Alg}(f(T)) + \Ocal \left( \tau_0 + \Delta \tau_{1, 2} \right)
    \\
    & \qquad + \Ocal \left( \frac{1}{\Delta} \left( \log T + d \log \log T + d \log \frac{d}{\Delta} \right)^2 \right)
    \, ,
\end{align*}
where in the last equality, the $\Ocal(\frac{d^2}{\Delta} \log^2\frac{d}{\Delta})$ term in the second term is absorbed into the last $\Ocal(\frac{1}{\Delta} \left( \log T + d \log \log T + d \log \frac{d}{\Delta} \right)^2 )$ term.
Therefore, there exists $\constterm{\tau_{\Alg}, f}$ and $G(T)$ such that $\constterm{\tau_{\Alg}, f} = \Ocal \left( \tau_0 + \Delta \tau_{1, 2} \right)$, $G(T) = \Ocal \left( \frac{1}{\Delta} \left( \log T + d \log \log T + d \log \frac{d}{\Delta} \right)^2 \right)$, and
\begin{align*}
    \Rcal_{\Greed{\Alg, \Tcal_e}}(T) 
    & \le \Rcal_{\Alg}(f(T)) + \constterm{\tau_{\Alg}, f} + G(T)
    \, .
\end{align*}
It remains to bound $\tau_0$ and $\tau_{1, 2}$.
Let $C_\beta > 0$ be a constant independent of $d, \Delta$, and $T$ that satisfies $\beta_T^2 \le C_\beta d \log (1 + T) $ for all $T$, which exists by \cref{lma:bound on alpha}.
Let $\tau_{1, 2}'$ be the least time step such that for all $T \ge \tau_{1, 2}'$, it holds that
\begin{align}
    \frac{32 C_\beta d \log(1 + 2T)}{\Delta^2} \sum_{t=1}^T \frac{1}{\max\{ f(t), 1\}} \le T
    \, .
    \label{eq:definition of tau12}
\end{align}
We show that $\tau_{1, 2} \le \max\{ \tau_0, \tau_{1, 2}'\}$.
For all $T \ge \tau_0 + \max\{ \tau_0, \tau_{1, 2}'\}$, it holds that
\begin{align*}
    \frac{8}{\Delta} \sum_{t \in \Gcal(\tau_0, T)} \frac{\beta_t^2}{f(t)}
    & \le 
    \frac{8 \beta_T^2}{\Delta} \sum_{t = \tau_0 + 1}^T \frac{1}{f(t)}
    \\
    & \le 
    \frac{8 \beta_T^2}{\Delta} \sum_{t = 1}^{T-\tau_0} \frac{1}{\max\{f(t), 1\}}
    \\
    & \le 
    \frac{8 C_\beta d \log (1 + T)}{\Delta} \sum_{t = 1}^{T-\tau_0} \frac{1}{\max\{f(t), 1\}}
    \\
    & \le 
    \frac{8 C_\beta d \log (1 + 2(T- \tau_0))}{\Delta} \sum_{t = 1}^{T-\tau_0} \frac{1}{\max\{f(t), 1\}}
    \\
    & \le \frac{1}{4} \Delta (T - \tau_0)
    \, ,
\end{align*}
where the first inequality holds since $\beta_t$ is increasing, the second inequality uses that $f(t)$ is increasing and $f(t) \ge 1$ for $t \ge \tau_0 + 1$, the third inequality holds by the definition of $C_\beta$, and the fourth inequality is due to $T \ge 2 \tau_0$, and the last inequality holds by the definition of $\tau_{1, 2}'$.
Therefore, we deduce that $\tau_{1, 2} \le \max\{ \tau_0, \tau_{1, 2}'\}$.
Then, we have that $\consttermdefault = \Ocal(\tau_0 + \Delta \tau_{1, 2} ) = \Ocal(\tau_0 + \Delta \tau_{1, 2}')$.

For some example functions $f$, we provide bounds on $\consttermdefault$ by providing bounds on $\tau_0$ and $\tau_{1, 2}'$.
We write $f(t) = \Omega(g(t))$ for a function $g(t)$ when there exist constants $C_1, C_2 > 0$ such that $f(t) \ge C_1 g(t) - C_2$ for all $t \in \NN$.

\begin{example}
\label{exm:linear}
Suppose $f(t) = \lfloor t / m \rfloor$ for some $m \in \NN$.
This case corresponds to executing $\Alg$ with a fixed period of $m$.
We have $f^{-1}(n) = m n$, so $\tau_0 = m \tau_{\Alg}$.
We now establish a bound on $\tau_{1, 2}'$ that satisfies Eq.~\eqref{eq:definition of tau12}.
We have
\begin{align*}
    \sum_{t=1}^T \frac{1}{\max\{ f(t), 1\}}
    & \le m + 
    \sum_{t=m+1}^T \frac{m}{t - m}
    \\
    & \le m (1 + \log T)
    \, .
\end{align*}
Using elementary analysis, one can show that after some time step $\tau = \Ocal(\frac{md}{\Delta^2} \log^2 \frac{md}{\Delta})$, it holds that $\frac{32 C_\beta m d }{\Delta^2} (1 + \log T)\log (1 + 2T) \le T $ for all $T \ge \tau$, hence $\tau_{1, 2}' = \Ocal(\frac{md}{\Delta^2} \log^2 \frac{md}{\Delta})$ holds.
Combining the bounds on $\tau_0$ and $\tau_{1, 2}'$, we obtain
\begin{align*}
    \consttermdefault = \Ocal \left( m \tau_{\Alg} + \frac{md}{\Delta} \log^2 \frac{md}{\Delta} \right)
    \, .
\end{align*}    
\end{example}

\begin{example}
\label{exm:log period}
Suppose $f(t) = \Omega(t / (\log t)^r)$ for some constant $r \ge 0$.
Then, $f^{-1}(n) = \Ocal( n (\log n)^r)$.
Also, we have
\begin{align*}
    \sum_{t=1}^T \frac{1}{\max\{f(t), 1\}}
    & = 
    \sum_{t=1}^T \Ocal \left( \frac{(\log t)^r}{t} \right)
    \\
    & = \Ocal \left( ( \log T)^{r+1} \right)
    \, .
\end{align*}
$\tau_{1, 2}'$ is the first time step such that $\Ocal(\frac{d}{\Delta^2} (\log T)^{r+2}) \le T$ for all $T \ge \tau_{1, 2}'$, and we can derive that $\tau_{1, 2}' = \Ocal( \frac{d}{\Delta^2} (\log \frac{d}{\Delta})^{r+2})$.
Therefore, we conclude that
\begin{align*}
    \consttermdefault = \Ocal \left( \tau_{\Alg} \left( \log \tau_{\Alg} \right)^r + \frac{d}{\Delta} \left( \log \frac{d}{\Delta} \right)^{r+2} \right)
    \, .
\end{align*}
    
\end{example}

\begin{example}
    Let $f(t) = \Omega(t^r)$ for some constant $r \in (0, 1)$.
    Then, $f^{-1}(n) = \Ocal(n^{1/r})$.
    We have
    \begin{align*}
        \sum_{t=1}^T \frac{1}{\max\{f(t), 1\}}
        & \le \sum_{t=1}^T \Ocal\left( \frac{1}{t^r} \right)
        \\
        &  = \Ocal( T^{1 - r})
        \, .
    \end{align*}
    For a constant $C > 0$, $ CT^{1 - r} \log T \le T$ is equivalent to $(C \log T)^{1/r} \le T$, and this inequality holds for all $T  \ge \tau$ with $\tau = \Ocal( (C \log C)^{1/r})$.
    Therefore, we have that for $\tau_{1, 2}' = \Ocal ( (\frac{d}{\Delta^2}\log \frac{d}{\Delta} )^{1/r} )$, it holds that $\Ocal( \frac{d}{\Delta^2} T^{1-r} \log T) \le T$ for all $T \ge \tau_{1, 2}'$.
    Therefore, we conclude that
    \begin{align*}
        \consttermdefault = \Ocal\left( \tau_{\Alg}^{\frac{1}{r}} + \frac{1}{\Delta^{\frac{2}{r} - 1}} \left( d \log \frac{d}{\Delta} \right)^{\frac{1}{r}} \right)
        \, .
    \end{align*}
\end{example}

\begin{example}
    Let $f(t) = \Omega( (\log t)^r)$ for some constant $r > 1$.
    Then, $f^{-1}(n) = e^{\Ocal( n^{1/r})}$.
    We have
    \begin{align*}
        \sum_{t=1}^T \frac{1}{\max\{f(t), 1\}}
        &
        = \sum_{t=1}^T \Ocal \left( \frac{1}{(\log t)^r} \right)
        \\
        & = \Ocal \left( \frac{T}{(\log T)^r} \right)
        \, .
    \end{align*}
    Then, $\tau_{1, 2}'$ must satisfy $\frac{C d T}{\Delta^2 (\log T)^{r-1}} \le T$ for some constant $C > 0$, or equivalently, $\frac{C d}{\Delta^2 } \le (\log T)^{r-1}$.
    We see that $\tau_{1, 2}' = \exp \left( \Ocal( ( d / \Delta^2 )^{1/(r-1)} ) \right)$.
    Therefore, we conclude that
    \begin{align*}
        \consttermdefault = \exp \left( \Ocal \left( \tau_{\Alg}^{\frac{1}{r}} \right) \right) + \Delta \exp \left( \Ocal \left( \left( d / \Delta^2 \right)^{\frac{1}{r-1}} \right) \right)
        \, .
    \end{align*}
\end{example}

\subsection{Proof of Theorem~\ref{thm:lower bound}}
\label{appx:lower bound}

\begin{proof}[Proof of Theorem~\ref{thm:lower bound}]
    For simplicity, we write $\pi := \Greed{\Alg, \Tcal_e}$.
    We analyze the performance of $\pi$ under two linear bandit instances.
    Let $\Delta > 0$ be a fixed constant whose value is chosen later.
    We define the arm set as $\Xcal = \{ e_1, \boldsymbol{0}_d \}$, where $e_1 \in \RR^d$ is the first standard basis vector and $\boldsymbol{0}_d \in \RR^d$ is the zero vector.
    This instance can be viewed as the one-armed bandit setting since the agent is aware that the second arm has reward $0$.
    The true parameter vectors are defined as $\theta_1 = (-\Delta, 0, \ldots, 0)$ and $\theta_2 = (\Delta, 0, \ldots, 0)$.
    In the first instance, $(\Xcal, \theta_1)$, the expected reward of the first arm is $-\Delta$, while the second arm yields a reward of $0$.
    Thus, the second arm is the optimal arm.
    Conversely, in the second instance, $(\Xcal, \theta_2)$, the first arm yields an expected reward of $\Delta$ and is the optimal arm.
    We assume that i.i.d. unit Gaussian noise is added to the observed reward.
    \\
    Fix $T \in \NN$.
    Let $N_1(T)$ and $N_2(T)$ be the number of times the first and second arms are selected up to time $T$, respectively.
    We define $\PP_1$ to be the probability distribution over the trajectory induced by policy $\pi$ interacting with instance $(\Xcal, \theta_1)$ for $T$ time steps, and define $\PP_2$ similarly for the second instance $(\Xcal, \theta_2)$.
    \\
    Let $D_{\text{KL}}(\cdot, \cdot)$ be the KL-divergence between two probability measures.
    By Lemma 15.1 in \citet{lattimore2020bandit}, we have that
    \begin{align*}
        D_{\text{KL}}(\PP_1, \PP_2) =  4 \Delta^2  \EE_1 [N_1(T) ] \, .
    \end{align*}
    Let $A := \{ N_1(T) < T / 2 \}$ be the event that the first arm is selected less than $T / 2$ times.
    By \cref{lma:TV to KL}, we obtain that
    \begin{align*}
        \PP_1 (A) + \PP_2 (A^{\mathsf{C}}) \ge \frac{1}{2} \exp ( - D_{\text{KL}}(\PP_1, \PP_2) )
        \, .
    \end{align*}
    Under the first instance, we have $\Rcal_\pi(T) = \Delta N_2(T)$.
    Using Markov's inequality, we obtain that $\EE_1 [ N_2(T) ] \ge \frac{T}{2} \PP_1( N_2(T) \ge \frac{T}{2}) = \frac{T}{2} \PP_1( N_1(T) < \frac{T}{2}) = \frac{T}{2} \PP_1(A) $, which implies that $\EE_1 [ \Rcal_\pi(T)]  \ge \frac{\Delta T}{2} \PP_1(A)$.
    Using a similar argument, we also derive that $\EE_2 [ \Rcal_\pi(T) ] \ge \frac{\Delta T}{2} \PP_2(A^{\mathsf{C}})$.
    Combining everything, we conclude that
    \begin{align}
        \EE_1 [ \Rcal_\pi (T) ] + \EE_2 [ \Rcal_\pi (T) ] 
        & \ge \frac{\Delta T}{2} \left( \PP_1 ( A) + \PP_2 (A^{\mathsf{C}}) \right)
        \nonumber 
        \\
        & \ge \frac{\Delta T}{4} \exp ( - D_{\text{KL}}(\PP_1, \PP_2))
        \nonumber 
        \\
        & = \frac{\Delta T}{4} \exp ( - 4 \Delta^2 \EE_1 [ N_1 (T) ])
        \label{eq:information lower bound}
        \, .
    \end{align}
    Now, we show that $\EE_1[ N_1(T) ]$ increases too slowly when $f(t) \ne \omega(\log t)$.
    First, we show that the expected number of greedy selections of the first arm under the first instance is at most a constant.
    Let $\hat{\mu}_1(T)$ be the empirical mean of the first arm after $T$ time steps.
    The greedy selection chooses the first arm only if $\hat{\mu}_1(T) \ge 0$.
    We bound the expected number of the averages of a Gaussian random walk exceeding $\Delta$ by the following lemma, whose proof is provided in \cref{appx:proof of random walk}:

    \begin{lemma}
    \label{lma:random walk}
        Let $Z_1, Z_2, \ldots$ be a sequence of i.i.d. samples of the unit Gaussian distribution and $S_n = \sum_{t=1}^n Z_t$ be its partial sum.
        Then, for any constant $c > 0$, the expected number of indices $n$ such that $S_n / n$ exceeds $c$ is at most $\frac{1}{2c^2}$, that is, $\EE [ \sum_{t=1}^\infty \ind\{ \frac{S_n}{n} \ge c \} ] \le \frac{1}{2c^2}$.
    \end{lemma}

    For $\hat{\mu}_1(T) \ge 0$ to hold, the average of the noises added to the random rewards of the first arm must be greater than $\Delta$. 
    Using \cref{lma:random walk}, we infer that
    \begin{align*}
        \EE_1 \left[ \sum_{t=1}^\infty \ind\{ X_t = e_1, \hat{\mu}_1(T) \ge 0 \}\right] \le \frac{1} {2\Delta^2}.
    \end{align*}
    Therefore, the expected number of suboptimal greedy selections is at most $\frac{1}{2\Delta^2}$.
    \\    
    Therefore, we have $\EE[N_1(T)] \le \frac{1}{2\Delta^2} + f(T)$ since there are at most $\frac{1}{2\Delta^2}$ suboptimal greedy selections and $f(T)$ exploratory selections.
    By $f(t) \ne \omega(\log t)$, there exists a constant $C > 0$ and infinitely many $T \in \NN$ such that $f(T) \le C \log T$.
    We conclude that for infinitely many $T$, we have $\EE[ N_1(T) ] \le \frac{1}{2\Delta^2} + C \log T$.
    Plugging this bound into Eq.~\eqref{eq:information lower bound}, we obtain that for infinitely many $T \in \NN$, it holds that
    \begin{align*}
        \EE_1[ \Rcal_\pi (T)] + \EE_2[ \Rcal_\pi (T)] 
        & \ge \frac{\Delta T}{4} \exp \left( - 4 \Delta^2 \left(\frac{1}{2\Delta^2} + C \log T \right) \right)
        \\
        & = \frac{\Delta}{4 e^2} T^{1 - 4\Delta^2 C}
        \, .        
    \end{align*}
    It implies that either $\EE_1[ \Rcal_\pi (T)]$ or $\EE_2[ \Rcal_\pi (T)] $ exceeds $\frac{\Delta}{8 e^2} T^{1 - 4\Delta^2 C}$.
    The proof is completed by taking $\Delta = \sqrt{\varepsilon / 4C}$ and $c(f, \varepsilon) = \frac{\Delta}{8e^2}$.
\end{proof}

\begin{remark}
    In the proof of \cref{thm:lower bound}, we show that $\EE_1 [N_1(T)] \le \left( \frac{1}{2\Delta^2} + C\log T \right)$ and $\EE_1 [\Rcal_\pi(T)] = \Delta \EE_1 [N_1(T)]$, so $\AlgName$ attains polylogarithmic regret for the first instance.
    Therefore, we can conclude that the instance that $\AlgName$ incurs almost linear regret is the second instance $(\Xcal, \theta_2)$.
\end{remark}

\section{Instance-Dependent Regret Analysis of Linear Thompson Sampling}
\label{appx:proof lints}

In this section, we provide an instance-dependent polylogarithmic regret bound of $\texttt{LinTS}$~\citep{agrawal2013thompson,abeille2017Linear}.
For completeness, we present the algorithm in~\cref{alg:LinTS}, where we use the version by~\citet{abeille2017Linear}.

\begin{algorithm}[t!] 
    \caption{Linear Thompson Sampling} 
    \begin{algorithmic}[1]
    \State Input : Sampling distribution $\Dcal^{\text{TS}}$
    \State Initialize $V_0 = I_d$
    \For{$t = 1, 2, \ldots, T$}
        \State Compute ridge estimator $\hat{\theta}_{t-1} = V_{t-1}^{-1} \sum_{i=1}^{t-1} X_i Y_i$
        \State Sample $\tilde{\eta}_t \sim \Dcal^{\text{TS}}$
        \State Compute perturbed parameter $\tilde{\theta}_t = \hat{\theta}_{t-1} + \beta_{t-1} V_{t-1}^{-1/2} \tilde{\eta}_t$
        \State Choose $X_t = \argmax_{x \in \Xcal} x^\top \tilde{\theta}_t$ and observe $Y_t$
        \State Update $V_t = V_{t-1} + X_t X_t^\top$
    \EndFor
    \end{algorithmic}
    \label{alg:LinTS}
\end{algorithm}

The input of the algorithm, $\Dcal^{\text{TS}}$, is a distribution over $\RR^d$.
We pose two conditions on the sampling distribution as in~\citet{abeille2017Linear}.

\begin{enumerate}
    \item (anti-concentration) There exists a positive probability $p$ such that for any $u \in \RR^d$ with $\| u \|_2 = 1$, 
    \begin{align*}
        \PP_{\eta \sim \Dcal^{\text{TS}}} (u^\top \eta \ge 1) \ge p
        \, .
    \end{align*}

    \item (concentration) There exists positive constants $c, c'$ such that for all $u \in \RR^d$ with $\| u \|_2 = 1$ and $\delta \in (0, 1]$, 
    \begin{align*}
        \PP_{\eta \sim \Dcal^{\text{TS}}} \left(  | u ^\top \eta | \le \sqrt{c \log \frac{c'}{\delta}} \right)) \ge 1 - \delta
        \, .
    \end{align*}
\end{enumerate}

The first condition comes directly from~\citet{abeille2017Linear}.
We slightly strengthen the second condition to derive a tighter bound when $\log  K \ll d$.
The original condition in~\citet{abeille2017Linear} poses that $\PP_{\eta \sim \Dcal^{\text{TS}}} \left(  \| \eta \|_2 \le \sqrt{cd \log (c'd / \delta)} \right)) \ge 1 - \delta$.
Our strengthened condition implies the original condition by taking $u$ to be the vectors of the standard basis and taking the union bound.
The strengthened condition holds for all the distributions discussed in~\citet{abeille2017Linear}, including the multivariate Gaussian distribution and spherical distribution.

Now, assuming that the conditions are true, we prove Theorem~\ref{thm:lints}.

\begin{proof}[Proof of Theorem~\ref{thm:lints}]
Let $\gamma_t := \beta_t(\delta) \min \big\{\sqrt{c d \log (2 c' d t^2 / \delta)}, \sqrt{c \log ( 2 c' K t^2 / \delta)} \big\}$.
Our choice of $\gamma_t$ slightly differs from~\citet{abeille2017Linear}; they choose it to be the first term in the minimum instead of taking the minimum over the two values.
We show that their analysis still applies even with this refined value of $\gamma_t$.
Suppose $\gamma_t = \beta_t(\delta) \sqrt{c \log ( 2 c' K t^2 / \delta)}$.
By the concentration condition on $\Dcal^{\text{TS}}$, for any $x \in \RR^d$, it holds that
\begin{align*}
    \PP_{t-1} \left( x^\top ( \beta_{t-1}(\delta) V_{t-1}^{-1/2} \tilde{\eta}_t) \le \beta_{t-1}(\delta) \| x \|_{V_{t-1}^{-1/2}}\sqrt{c \log ( 2 c' t^2 / \delta) } \right) \ge 1 - \frac{\delta}{2 t^2}
    \, .
\end{align*}
Taking the union bound over $x \in \Xcal$, we obtain
\begin{align*}
    &\PP_{t-1} \left( \forall x \in \Xcal, x^\top ( \beta_{t-1}(\delta) V_{t-1}^{-1/2} \tilde{\eta}_t) \le  \beta_{t-1}(\delta) \| x \|_{V_{t-1}^{-1/2}}\sqrt{c \log ( 2 c' K t^2 / \delta) } \right)
    \\
    & = 
    \PP_{t-1} \left( \forall x \in \Xcal, x^\top ( \beta_{t-1}(\delta) V_{t-1}^{-1/2} \tilde{\eta}_t) \le  \gamma_t \| x \|_{V_{t-1}^{-1/2}} \right)
    \\
    & \ge 1 - \frac{\delta}{2 t^2}
    \, .
\end{align*}
This probabilistic inequality is the only property $\gamma_t$ must satisfy in the analysis of~\citet{abeille2017Linear}, therefore the results in their paper hold for this refined value of $\gamma_t$.

We first decompose the instantaneous regret of \texttt{LinTS} as follows:

\begin{align*}
    \text{reg}_t & = x^{*\top}\theta^* - X_t^\top \theta^*
    \\
    & = \underbrace{x^{*\top} \theta^* - X_t^\top \tilde{\theta}_{t-1}}_{R_t^{\text{TS}}} + \underbrace{X_t^\top \tilde{\theta}_{t-1} - X_t^\top \theta^*}_{R_t^{\text{RLS}}}
    \, .
\end{align*}

Following the proof of~\citet{abeille2017Linear}, we obtain that $R_t^{\text{TS}} \le \frac{4 \gamma_t}{p} \EE_{t-1} \left[ \| X_t \|_{V_{t-1}^{-1}} \right]$ and $R_t^{\text{RLS}} \le \beta_t(\delta)  \| X_t \|_{V_{t-1}^{-1}}$.
By the definition of the minimum gap $\Delta$, we have either $\text{reg}_t = 0$ or $\text{reg}_t \ge \Delta$, which implies that $\text{reg}_t \le \frac{\text{reg}_t^2}{\Delta}$.
Therefore, we derive the following bound on $\reg_t$.
\begin{align*}
    \reg_t
    & \le \frac{\reg_t^2}{\Delta}
    \\
    & = \frac{ ( R_t^{\text{TS}} + R_t^{\text{RLS}})^2}{\Delta}
    \\
    & \le \frac{ 2 (R_t^{\text{TS}})^2 +2 ( R_t^{\text{RLS}})^2}{\Delta}
    \\
    & \le \frac{2}{\Delta} \left( \frac{16 \gamma_t^2}{p^2} \EE_{t-1} \left[ \| X_t \|_{V_{t-1}^{-1}} \right]^2 + \beta_t(\delta) ^2 \| X_t \|_{V_{t-1}^{-1}}^2 \right)
    \\
    & \le \frac{2}{\Delta} \left( \frac{16 \gamma_t^2}{p^2} \EE_{t-1} \left[ \| X_t \|_{V_{t-1}^{-1}}^2 \right] + \beta_t(\delta) ^2 \| X_t \|_{V_{t-1}^{-1}}^2 \right)
    \, ,
\end{align*}
where the second inequality uses that $(a + b)^2 \le 2a^2 + 2 b^2$ for all $a, b \in \RR$, and the last inequality is due to Jensen's inequality.
We bound $\sum_{t=1}^T \EE_{t-1} [ \| X_t \|_{V_{t-1}^{-1}}^2 ]$ using the following lemma that provides a lower bound for a sum of nonnegative random variables.
Its proof is provided in \cref{appx:proof of freedman app}.

\begin{lemma}
\label{lma:freedman app}
    Let $\{X_t \}_{t=1}^\infty$ be a sequence of real-valued random variables adapted to a filtration $\{ \Fcal_t \}_{t=0}^\infty$.
    Suppose $0 \le X_t \le 1$ for all $t$.
    For any $\delta \in (0, 1]$, the following inequality holds for all $n \in \NN$ with probability at least $1 - \delta$:
    \begin{align*}
        \sum_{t=1}^n \EE [ X_t \mid \Fcal_{t-1} ] \le 2 \sum_{t=1}^n X_t + 2 \log \frac{1}{\delta}
        \, .
    \end{align*}
\end{lemma}

Applying \cref{lma:freedman app} on $\{ \| X_t \|_{V_{t-1}^{-1}} ^2\}_t$, we derive that with probability at least $1 - \delta$, it holds that
\begin{align*}
    \sum_{t=1}^T \EE_{t-1} \left[  \| X_t \|_{V_{t-1}^{-1}}^2 \right] \le 2 \sum_{t=1}^T  \| X_t \|_{V_{t-1}^{-1} }^2+ 2 \log \frac{1}{\delta}
    \, .
\end{align*}
for all $T \in \NN$.
Therefore, the cumulative regret of $\texttt{LinTS}$ is bounded as follows:
\begin{align*}
    \Rcal_{\texttt{LinTS}}(T)
    & \le \sum_{t=1}^T \frac{2}{\Delta} \left( \frac{16 \gamma_t^2}{p^2} \EE_{t-1} \left[ \| X_t \|_{V_{t-1}^{-1}}^2 \right] + \beta_t(\delta) ^2 \| X_t \|_{V_{t-1}^{-1}}^2 \right)
    \\
    & \le \frac{2}{\Delta} \left( \left(\frac{32 \gamma_T^2}{p^2} + \beta_T(\delta) ^2 \right) \sum_{t=1}^T \| X_t \|_{V_{t-1}^{-1}}^2 + \frac{32 \gamma_T^2}{p^2} \log \frac{1}{\delta} \right)
    \\
    & \le \frac{4}{\Delta} \left( \left(\frac{32 \gamma_T^2}{p^2} + \beta_T(\delta) ^2 \right) \alpha_T + \frac{16 \gamma_T^2}{p^2} \log \frac{1}{\delta} \right)
    \, ,
\end{align*}
where the third inequality applies \cref{lma:elliptical potential}.
Finally, plugging in $\beta_T(\delta) ^2 = \Ocal (\alpha_T + \log \frac{1}{\delta})$ and $\gamma_T^2 = \Ocal( \min\{ d\log \frac{dT}{\delta}, \log \frac{KT}{\delta} \} (\alpha_T + \log \frac{1}{\delta}))$ proves the theorem.
\end{proof}

\section{Proofs of Technical Lemmas}
\label{appx:proof lemma}

In this section, we provide proofs of Lemmas~\ref{lma:large enough t}~to~\ref{lma:concentration of optimal arm}.

\subsection{Proof of Lemma~\ref{lma:large enough t}}
\label{appx:proof of large enough t}

\begin{proof}[Proof of Lemma~\ref{lma:large enough t}]
    Take $\tau_{\Alg'}$ to be the least positive integer that satisfies
    \begin{align*}
        \frac{C d^a}{\Delta^b} \log^c T \le \frac{3}{4} \Delta T
    \end{align*}
    for all $T \ge \tau_{\Alg'}$, which exists since $\displaystyle \lim_{T \rightarrow \infty} \frac{\log^c T}{T}  = 0$.
    Elementary analysis shows that $\tau_{\Alg'} = \Ocal( \frac{d^a}{\Delta^{b+1}} \log ^c \frac{d}{\Delta})$.
    Let $\Nsub(T)$ be the number of suboptimal selections made by $\Alg'$ up to time step $T$.
    Since a suboptimal selection incurs at least $\Delta$ regret, we have $\Delta \Nsub(T) \le \Rcal_{\Alg'}(T)$.
    It implies that for any $ T \ge \tau_{\Alg'}$, we have $\Delta \Nsub(T) \le \frac{3}{4} \Delta T$, or equivalently, $\Nopt(T) \ge \frac{1}{4} T$, which proves the lemma.
\end{proof}

\subsection{Proof of Lemma~\ref{lma:first decomposition}}

In this subsection, we prove Lemma~\ref{lma:first decomposition}.
To do so, we show that the estimation error of the optimal reward $x^{*\top} \theta^*$ scales with $\frac{1}{\sqrt{\Nopt(t)}}$, where we need the following technical lemma.
Its proof is deferred to \cref{appx:proof of Nopt lemma}.

\begin{lemma}
\label{lma:concentration of optimal arm}
    We have that for all $t \in \NN$,
    \begin{align*}
        \| x^* \|_{V_{t}^{-1}}^2 \le \frac{1}{1 + \Nopt(t)}
        \, .
    \end{align*}
\end{lemma}

Now, we prove \cref{lma:first decomposition}.

\begin{proof}[Proof of Lemma~\ref{lma:first decomposition}]
    The instantaneous regret of a greedy selection can be bounded as follows:
    \begin{align*}
        \text{reg}_t
        & = x^{*\top} \theta^* - X_t^{\top} \theta^*
        \\
        & \le x^{*\top} \theta^* - x^{*\top} \hat{\theta}_{t-1} + X_t^\top \hat{\theta}_{t-1} - X_t^{\top} \theta^*
        \\
        & = x^{*\top} \left( \theta^* - \hat{\theta}_{t-1} \right) + X_t^\top \left( \hat{\theta}_{t-1} - \theta^* \right)
        \\
        & \le \left( \| x^* \|_{V_{t-1}^{-1}} + \| X_t \|_{V_{t-1}^{-1}} \right) \| \theta^* - \hat{\theta}_{t-1} \|_{V_{t-1}}
        \\
        & \le \beta_{t-1} \left( \| x^* \|_{V_{t-1}^{-1}} + \| X_t \|_{V_{t-1}^{-1}} \right)
        \, ,
    \end{align*}
    where the first inequality uses that $x^{*\top} \hat{\theta}_{t-1} \le X_t^\top \hat{\theta}_{t-1}$ when $X_t$ is chosen greedily, the second inequality is due to the Cauchy-Schwarz inequality, and the last inequality comes from \cref{lma:self normalization}.
    By the definition of the minimum gap $\Delta$, we have either $\text{reg}_t = 0$ or $\text{reg}_t \ge \Delta$, which implies that $\text{reg}_t \le \frac{\text{reg}_t^2}{\Delta}$.
    Then, we obtain that
    \begin{align*}
        \text{reg}_t
        & \le \frac{\text{reg}_t^2}{\Delta}
        \\
        & \le \frac{\beta_{t-1}^2 \left( \| x^* \|_{V_{t-1}^{-1}} + \| X_t \|_{V_{t-1}^{-1}}\right)^2 }{\Delta}
        \\
        & \le \frac{2 \beta_{t-1}^2 \left( \| x^* \|_{V_{t-1}^{-1}}^2 + \| X_t \|_{V_{t-1}^{-1}}^2\right)}{\Delta}
        \, ,
    \end{align*}
    where the last inequality uses that $(a+b)^2 \le 2 (a^2 + b^2)$ for any $a, b \in \RR$.
    Taking the sum of instantaneous regret for $t \in \Gcal(\tau, T)$, we proceed as follows:
    \begin{align*}
        \Rcal_{\Greed{\Alg, \Tcal_e}}^G(\tau, T)
        & = \sum_{t \in \Gcal(\tau, T)} \text{reg}_t
        \\
        & \le \sum_{t \in \Gcal(\tau, T)} \frac{2 \beta_{t-1}^2 \left( \| x^* \|_{V_{t-1}^{-1}}^2 + \| X_t \|_{V_{t-1}^{-1}}^2\right)}{\Delta}
        \\
        & \le \frac{2 \beta_T^2}{\Delta} \sum_{t \in \Gcal(\tau, T)} \| X_t \|_{V_{t-1}^{-1}}^2 + \frac{2 }{\Delta} \sum_{t \in \Gcal(\tau, T)} \beta_{t-1}^2 \| x^* \|_{V_{t-1}^{-1}}^2
        \\
        & \le \frac{4 \alpha_T \beta_T^2}{\Delta} + \frac{2 }{\Delta} \sum_{t \in \Gcal(\tau, T)} \beta_{t-1}^2 \| x^* \|_{V_{t-1}^{-1}}^2
        \\
        & \le \frac{4 \alpha_T \beta_T^2}{\Delta} + \frac{2 }{\Delta} \sum_{t \in \Gcal(\tau, T)} \frac{\beta_{t-1}^2}{1 + \Nopt(t-1)}
        \, ,
    \end{align*}
    where the third inequality is due to \cref{lma:elliptical potential} and the last inequality applies \cref{lma:concentration of optimal arm}.
\end{proof}

\subsection{Proof of Lemma~\ref{prop:weak regret bound}}

\begin{proof}[Proof of Lemma~\ref{prop:weak regret bound} ]
    By the choice of $\tau_{\Alg}$, at least a quarter of the selections by $\Alg$ are optimal when $f(t) \ge \tau_{\Alg}$, or equivalently, $t \ge f^{-1}(\tau_{\Alg})$.
    It implies that $\Nopt(t) \ge \frac{1}{4} f(t) $.
    Then, it holds that $1 + \Nopt(t-1) \ge 1 + \frac{1}{4}f(t-1) \ge 1 + \frac{1}{4}(f(t) - 1) \ge \frac{1}{4}f(t)$.
    Plugging this bound into \cref{lma:first decomposition}, we conclude that
    \begin{align*}
        \Rcal_{\Greed{\Alg, \Tcal_e}}^G ( f^{-1}(\tau_{\Alg}, T)
        & \le \frac{4 \alpha_T \beta_T^2}{\Delta} + \frac{2}{\Delta}\sum_{t \in \Gcal(f^{-1}(\tau_{\Alg}), T)} \frac{\beta_{t-1}^2}{1 + \Nopt(t-1)}
        \\
        & \le \frac{4 \alpha_T \beta_T^2}{\Delta} + \frac{8}{\Delta}\sum_{t \in \Gcal(f^{-1}(\tau_{\Alg}), T)} \frac{\beta_{t-1}^2}{f(t)}
        \\
        & \le \frac{4 \alpha_T \beta_T^2}{\Delta} + \frac{8}{\Delta}\sum_{t \in \Gcal(f^{-1}(\tau_{\Alg}), T)} \frac{\beta_{t}^2}{f(t)}
        \, .
    \end{align*}
    Now, we show that this quantity is sublinear in $T$.
    By \cref{lma:bound on alpha}, we have $\alpha_T, \beta_T^2 = \Ocal(d\log T)$, so $\frac{4\alpha_T \beta_T^2}{\Delta}$ is sublinear in $T$.
    By $f(t) = \omega(\log t)$ and $\beta_t^2 = \Ocal(d \log T)$, we have $\lim_{t \rightarrow \infty} \frac{\beta_t^2}{f(t)} = 0$, which implies that $\sum_{t \in \Gcal(f^{-1}(\tau_{\Alg}), T)} \frac{\beta_{t}^2}{f(t)}$ is sublinear in $T$.    
\end{proof}

\subsection{Proof of Lemma~\ref{lma:better bound on alpha}}

\begin{proof}[Proof of Lemma~\ref{lma:better bound on alpha}]

    We decompose $V_T$ as follows:
    \begin{align*}
        V_{T}
        & = I_d + \sum_{t=1}^T X_t X_t^\top
        \\
        & =  I_d  + \sum_{t = 1}^T \ind\{ X_t = x^*\} X_t X_t^{\top} + \sum_{t = 1}^T \ind\{ X_t \ne x^*\} X_t X_t^{\top}
        \\
        & =  I_d + \Nopt(T) x^* x^{*\top} + \sum_{t = 1}^T \ind\{ X_t \ne x^*\} X_t X_t^{\top}
        \\
        & = : A + B
        \, ,
    \end{align*}
    where we define $A:=  I_d +\Nopt(T) x^* x^{*\top}$ and $B:= \sum_{t = 1}^T \ind\{ X_t \ne x^*\} X_t X_t^{\top}$.
    The eigenvalues of $A$ are $1+ \Nopt(T) \| x^* \|_2, 1, \ldots, 1$.
    Let $b_1 \ge b_2 \ge \ldots \ge b_d$ be the eigenvalues of $B$.
    Finally, let $v_1 \ge v_2 \ge \ldots \ge v_d$ be the eigenvalues of $V_T$.
    By \cref{lma:Weyl inequality}, we have
    \begin{align*}
        v_1 \le (1 + \Nopt(T) \| x^* \|_2) + b_1
    \end{align*}
    and
    \begin{align*}
        v_i \le \lambda_2(A) + b_{i-1} = 1 + b_{i - 1} 
    \end{align*}
    for $i = 2, \ldots, d$.
    Let $\Nsub(T) := T - \Nopt(T)$ be the number of suboptimal arm selections up to time $T$.
    Then, we have $b_1 \le \tr(B) \le \Nsub(T)$, so we infer that
    \begin{align*}
        v_1 \le (1 + \Nopt(T) \| x^* \|_2) + b_1 & \le 1 + \Nopt(T) \| x^* \|_2 + \Nsub(T)
        \le 1 + T
        \, .
    \end{align*}
    and 
    \begin{align*}
        \Pi_{i=2}^d v_i
        & \le \Pi_{i=2}^d \left( 1 + b_{i-1} \right)
        \\
        & \le \left( \frac{\sum_{i=2}^d \left( 1 + b_{i-1} \right) }{d - 1} \right)^{d-1}
        \\
        & \le \left( 1 + \frac{\tr(B)}{d - 1} \right)^{d-1}
        \\
        & \le \left( 1 + \frac{\Nsub(T)}{d - 1} \right)^{d-1}
        \, ,
    \end{align*}
    where the second inequality is the AM-GM inequality.
    Then, we have
    \begin{align*}
        \alpha_T &  = \log \frac{\det V_T}{\det V_0}
        \\
        & = \sum_{i=1}^d \log v_i
        \\
        & \le \log (1 + T) + (d - 1) \log \left( 1 + \frac{\Nsub(T)}{(d - 1)} \right) 
        \, .
    \end{align*}
    Since a suboptimal selection incurs at least $\Delta$ regret, we have that $\Delta \Nsub(T) \le \Rcal_{\Alg'}(T)$, or equivalently, $\Nsub(T) \le \frac{1}{\Delta} \Rcal_{\Alg'}(T)$.
    Plugging in this bound completes the proof.
\end{proof}

\subsection{Proof of Lemma~\ref{lma:random walk}}
\label{appx:proof of random walk}

\begin{proof}[Proof of Lemma~\ref{lma:random walk}]
    Let $\Phi(\cdot)$ be the cumulative density function of the standard Gaussian distribution.
    Since the distribution of $S_n / n$ follows the Gaussian distribution with mean $0$ and variance $1 / n$, we have $\PP( \frac{S_n}{n} \ge c) = 1 - \Phi(c \sqrt{n})$.
    Then, we have that
    \begin{align*}
        \EE \left[ \sum_{n=1}^\infty \ind \left\{ \frac{S_n}{n} \ge c \right\} \right]
        & = \sum_{n=1}^\infty \EE \left[ \ind \left\{ \frac{S_n}{n} \ge c  \right\} \right]
        \\
        & =\sum_{n=1}^\infty (1 - \Phi(c \sqrt{n}))
        \, .
    \end{align*}
    Since $1 - \Phi(c\sqrt{n})$ is a decreasing function with respect to $n$, we can upper bound the summation by an integral and conclude as follows:
    \begin{align*}
        \sum_{n=1}^\infty (1 - \Phi(c \sqrt{n}))
        & \le \int_{0}^\infty 1 - \Phi(c \sqrt{t}) \, dt
        \\
        & = \int_0^\infty \int_{c\sqrt{t}}^\infty \frac{1}{\sqrt{2\pi}} e^{- \frac{x^2}{2}} \, dx \, dt
        \\
        & = \int_{0}^\infty \int_0^{\left( \frac{x}{c} \right)^2} \frac{1}{\sqrt{2\pi}} e^{- \frac{x^2}{2}} \, dt\, dx         
        \\
        & = \int_{0}^\infty \left( \frac{x}{c} \right)^2 \frac{1}{\sqrt{2\pi}} e^{- \frac{x^2}{2}} \, dx 
        \\
        & = \frac{1}{2c^2}
        \, ,
    \end{align*}
    where the first equality plugs in the probability density function of the Gaussian distribution and the second equality interchanges the order of the integral, which is justified by Fubini's theorem since the integrand is continuous and positive.
\end{proof}

\subsection{Proof of Lemma~\ref{lma:freedman app}}
\label{appx:proof of freedman app}
\begin{proof}[Proof of Lemma~\ref{lma:freedman app}]
    For simplicity, denote $\EE [ \cdot \mid \Fcal_{t-1} ]$ by $\EE_{t-1} [ \cdot ]$.
    By $e^x \le 1 + x + \frac{x^2}{2}$ for all $x \le 0$, we have that
    \begin{align*}
        \EE_{t-1} [ e^{ - X_t} ]
        & \le \EE_{t-1} [ 1 - X_t + \frac{1}{2} X_t^2]
        \\
        & = 1 - \EE_{t-1} [ X_t ] + \frac{1}{2} \EE_{t-1} [  X_t^2 ]
        \\
        & \le 1 - \frac{1}{2}\EE_{t-1} [ X_t ]
        \\
        & \le e^{- \frac{1}{2}\EE_{t-1} [ X_t ]}
        \, ,
    \end{align*}
    where the second inequality uses that $X_t \ge 0$ and $X_t^2 \le X_t$ when $0 \le X_t \le 1$ and the last inequality holds since $1 + x \le e^x$ for all $x \in \RR$.
    Then, $M_n := \exp \left( \sum_{t=1}^n \left( - X_t + \frac{1}{2} \EE_{t-1} [X_t ] \right) \right) $ is a supermartingale.
    By Ville's maximal inequality, we have that $\PP( \exists n \in \NN : M_n \ge \frac{1}{\delta} ) \le \delta$.
    Taking the logarithm and rearranging the terms leads to the following conclusion:
    \begin{align*}
        \PP \left( \exists n \in \NN : \sum_{t=1}^n \EE_{t-1} [ X_t ] \ge 2 \sum_{t=1}^n X_t + 2 \log \frac{1}{\delta} \right) \le \delta
        \, .
    \end{align*}
\end{proof}

\subsection{Proof of Lemma~\ref{lma:concentration of optimal arm}}
\label{appx:proof of Nopt lemma}

We prove Lemma~\ref{lma:concentration of optimal arm} by proving the following more general lemma.

\begin{lemma}
\label{lma:matrix norm lemma}
    For $\lambda, n > 0$ and $x \in \RR^d$, let $V$ be a symmetric matrix with $V \succeq \lambda I_d + n x x^\top$.
    Then, $\| x \|_{V^{-1}}^2 \le \frac{1}{\lambda + n}$.
\end{lemma}
\begin{proof}
    It is sufficient to consider the case $V = \lambda I_d + n x x^\top$ only since $\| x \|_{V^{-1}}^2 \le \| x \|_{(\lambda I_d + n x x^\top)^{-1}}^2$.
    In this case, we have
    \begin{align*}
        V x
        & = \lambda x + n x x^\top x
        \\
        & = \left( \lambda + n \|x \|_2^2 \right) x
        \, .
    \end{align*}
    Multiply $x^{\top}V^{-1}$ on the left to both sides and obtain
    \begin{align*}
        \| x \|_2^2 & = \left( \lambda + n \| x \|_2^2 \right) \| x \|_{V^{-1}}^2
        \, .
    \end{align*}
    By reordering the terms, we obtain that
    \begin{align*}
        \| x \|_{V^{-1}}^2
        & = \frac{ \| x \|_2^2}{\lambda + n \| x \|_2^2} 
        \le \frac{1}{\lambda + n}
        \, ,
    \end{align*}
    completing the proof.
\end{proof}

\section{Auxiliary Lemmas}

\label{appx:auxiliary}

Recall that $\alpha_T = \log \frac{ \det V_T}{\det V_0}$ and $\beta_T(\delta) = \sigma \sqrt{\alpha_T + 2 \log (1 / \delta)} + S$.

\begin{lemma}[Theorem 2 in \citet{abbasi2011improved}]
\label{lma:self normalization}
    With probability at least $1 - \delta$, ${\big\| \theta^* - \hat{\theta}_t \big \|_{V_t} \le \beta_t(\delta)}$ holds
    for all $t \ge 0$.
\end{lemma}

\begin{lemma}[Lemma 10 in \citet{abbasi2011improved}]
\label{lma:bound on alpha}
    It holds that $ \alpha_T \le d \log ( 1 + \frac{T}{d})$.
\end{lemma}

\begin{lemma}[Lemma 11 in \citet{abbasi2011improved}]
\label{lma:elliptical potential}
    For any sequence of $X_1, \ldots, X_T$ with $X_t \in \BB^d$ for all $t = 1, \ldots, T$, we have
    $\sum_{t=1}^T \| X_t \|_{V_{t-1}^{-1}}^2 \le 2 \alpha_T$.
\end{lemma}

\begin{lemma}[Bretagnolle-Huber inequality~\citep{bretagnolle1979estimation}, Theorem 14.2 in~\citet{lattimore2020bandit}]
\label{lma:TV to KL}
Let $\PP$ and $\QQ$ be two probability measures on the same measurable space $(\Omega, \Fcal)$.
Let $D_{\text{KL}}(\PP, \QQ) := \int \log \frac{d\PP}{d\QQ} d\PP$ be the Kullback-Leibler divergence between $\PP$ and $\QQ$.
Then, for any event $A \in \Fcal$, it holds that
\begin{align*}
    \PP( A) + \QQ(A^{\mathsf{C}}) \ge \frac{1}{2} \exp ( D_{\text{KL}}(\PP, \QQ) )
    \, .
\end{align*}

\begin{lemma}[Weyl's inequality \citep{weyl1912asymptotische}]
\label{lma:Weyl inequality}
For a Hermitian matrix $A \in \CC^{d \times d}$, let $\lambda_1(A) \ge \cdots \ge \lambda_d(A)$ be its eigenvalues sorted from large to small.
For two Hermitian matrices $A, B \in \CC^{d \times d}$ and any $1 \le i, j \le d$ with $i + j - 1 \le d$, it holds that
\begin{align*}
    \lambda_{i + j - 1} (A + B) \le \lambda_i (A) + \lambda_j (B)
    \, .
\end{align*}
\end{lemma}

\end{lemma}

\section{Extension to Time-Varying Features}
\label{appx:context}

In this section, we discuss the possibility of relaxing the assumption of requiring a finite and fixed arm set.

Previous literature on greedy bandit algorithms~\citep{bastani2021mostly,kannan2018smoothed,sivakumar2020structured,raghavan2023greedy,kim2024local} has established the effectiveness of purely greedy selections under certain favorable context distributions, specifically when features are drawn i.i.d. from distributions with suitable diversity conditions. Under such conditions, the regret contributions from the base exploratory algorithm and greedy selections can be analyzed separately. Moreover, since our analysis primarily assumes a fixed optimal arm $x^*$, the theoretical results provided in \cref{thm:main theorem} readily extend to contexts where the optimal arm remains invariant.

However, an important and open challenge remains: extending the performance guarantees of $\AlgName$ to scenarios involving dynamically varying optimal arms. Addressing these more general cases is non-trivial, as our current analysis relies on the property that estimation errors of $x^{*\top}\theta^*$ diminish when the optimal arm is selected frequently. This property becomes less straightforward to guarantee when the optimal arm itself is random or time-varying.
Notably, pointwise guarantees for linear regression with random design require additional distributional assumptions~\citep{hsu2012random}, suggesting that bounding the estimation error of a random optimal arm without assumptions may be infeasible.

Meanwhile, \citet{hanna2023contexts} propose a reduction technique that enables linear bandit algorithms to address linear contextual bandit problems when the arm set is sampled i.i.d. from a fixed distribution. Their results, however, focus on worst-case $\Ocal(\sqrt{T})$-type regret, which is suboptimal in our context where instance-dependent polylogarithmic regret is desired. Additionally, while a greedy selection chooses the same arm irrespective of this reduction, the parameter update involves a mismatch: the observed reward $Y_t$ from the selected arm $X_t$ is attributed to a potentially different predetermined vector $X_t'$. Despite these challenges, the approach by \citet{hanna2023contexts} underscores the feasibility of adapting linear bandit methods to contextual scenarios, suggesting promising directions for extending our results in future work.

\newpage
\section{Additional Experiments}
\label{appx:experiment}

\begin{figure}[!ht]
    \centering
    \includegraphics[width=\linewidth]{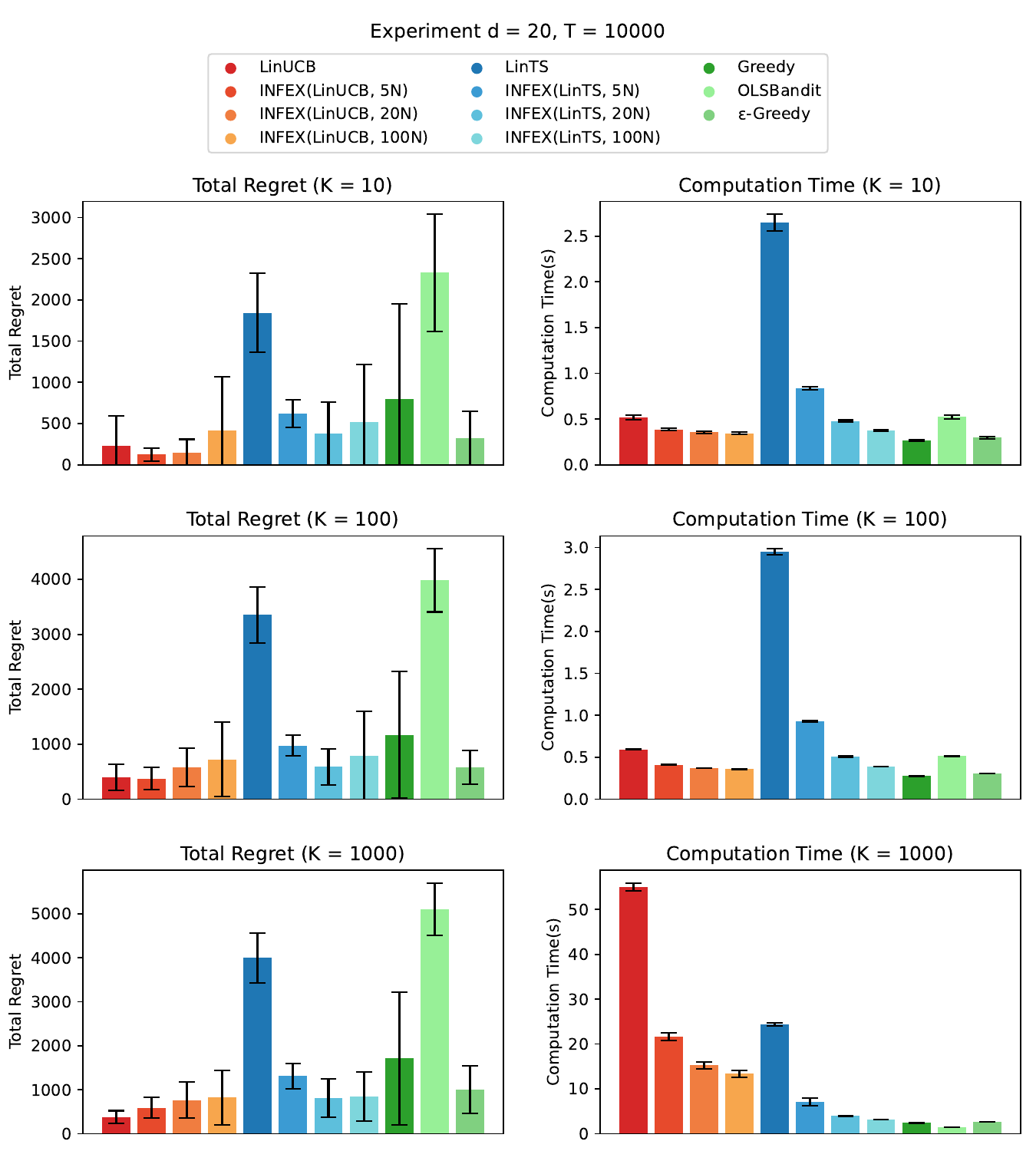}
    \caption{Comparison of total regret (left) and computation time (right) when $d = 20$, $T = 10000$, and $K = 10$ (top), $K = 100$ (middle), and $K = 1000$ (bottom).}
    \label{fig:d20}
\end{figure}

\begin{figure}[!ht]
    \centering
    \includegraphics[width=\linewidth]{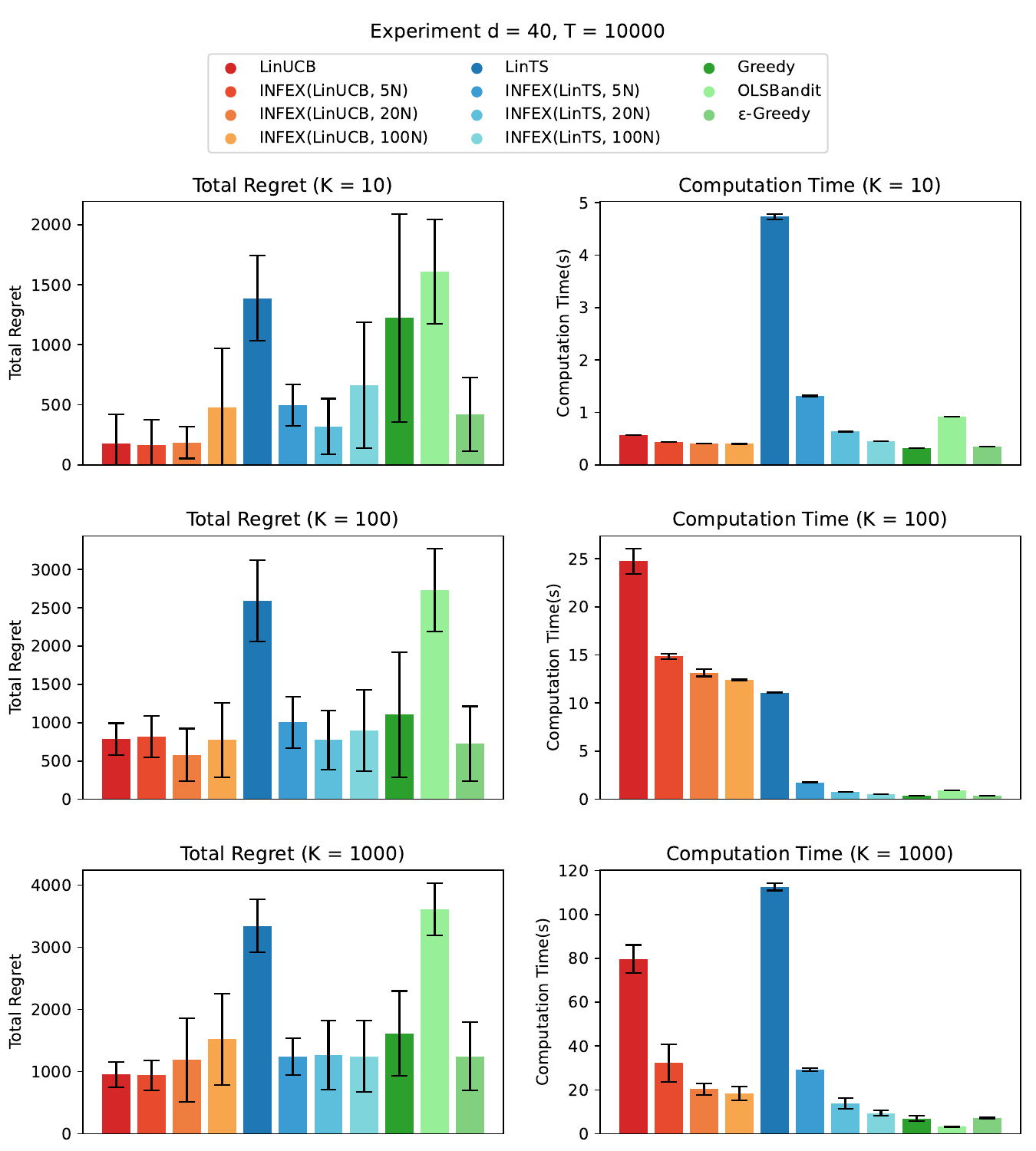}
    \caption{Comparison of total regret (left) and computation time (right) when $d = 40$, $T = 10000$, and $K = 10$ (top), $K = 100$ (middle), and $K = 1000$ (bottom).}
    \label{fig:d40}
\end{figure}

We provide additional experimental results for different values of $d$ omitted in \cref{sec:experiment}.
Except for the difference in the ambient dimension $d$, the generation of the problem instances and the algorithms is identical to those described in \cref{sec:experiment}.
Figure~\ref{fig:d20} presents the result when $d = 20$ and Figure~\ref{fig:d40} presents the result when $d = 40$.
We observe the same trends as in the case where $d = 10$.
Even for larger $d$, $\AlgName$ consistently demonstrates efficiency in both regret and computational time.

All hyperparameters of the algorithms are set to their theoretical values.
Both $\texttt{LinUCB}$ and $\texttt{LinTS}$ require the confidence radius $\beta_t$.
We explicitly compute the value of $\log \frac{\det V_t}{\det V_0}$ using rank-one update~\citep{abbasi2011improved} instead of using its upper bound $d \log T$, so that the base algorithms achieve the regret bounds of Theorem 5 in \citet{abbasi2011improved} and \cref{thm:lints}.
We expect that the regret reduction achieved by $\AlgName$ would have been even more significant if the base algorithm had used a crude upper bound for the confidence radius.

The experiments are conducted on a computing cluster with twenty Intel(R) Xeon(R) Silver 4214R CPUs, and three of them are used for the experiments.
The total runtime of the entire experiment is approximately one hour.

\end{document}